\documentclass[11pt]{article}
\usepackage[top=1.2in,bottom=1.2in,left=1.2in,right=1.2in]{geometry}
\usepackage[numbers]{natbib}

\usepackage{lineno,hyperref}
\usepackage{booktabs}       
\usepackage{graphicx}
\usepackage{times}
\usepackage{amsmath}
\usepackage{amssymb}
\usepackage{appendix}
\usepackage{rotating}
\usepackage{enumerate}
\usepackage{fancyhdr}
\usepackage{placeins}
\usepackage{caption}
\usepackage{subcaption}
\usepackage{multirow}
\usepackage{listings}
\usepackage{color}
\usepackage{pdflscape}
\usepackage{pgfgantt}
\usepackage{amsthm}
\usepackage{upgreek}
\usepackage{amsmath}
\usepackage{esvect}
\usepackage[ruled,noline,noend]{algorithm2e}
\usepackage{cleveref}       

\SetKwInput{KwInput}{Input}
\modulolinenumbers[5]

\newtheorem{definition}{Definition}

\newcommand{\model}[2]{\ensuremath{f^{#1}_{#2}}}
\newcommand{\modelset}{\ensuremath{\mathcal{M}}}
\newcommand{\modelsubset}{\ensuremath{{\mathcal{M}}'}}
\newcommand{\metamodel}[1]{\ensuremath{F^{#1}}}
\newcommand{\concept}[2]{\ensuremath{c^{#1}_{#2}}}

\newcommand{\datastream}[1]{\ensuremath{X^{#1}\!}}
\newcommand{\datastreamsegment}[2]{\ensuremath{X^{#1}_{#2}\!}}

\newcommand{\labelspace}[1]{\ensuremath{Y^{#1}\!}}
\newcommand{\labelsegment}[2]{\ensuremath{Y^{#1}_{#2}\!}}

\newcommand{\window}{\ensuremath{W}}
\newcommand{\pcs}[1]{\ensuremath{{U}_{#1}}}

\newcommand{\instance}[1]{\ensuremath{x_{#1}}}
\newcommand{\tLabel}[1]{\ensuremath{y_{#1}}}
\newcommand{\conceptSet}[1]{\ensuremath{C^{#1}}}
\newcommand{\predict}[1]{\ensuremath{\hat{y}_{#1}^{*}}}

\newcommand{\domain}[1]{\ensuremath{\mathcal{D}^{#1}\!}}

\newcommand{\realnumbers}[1]{\ensuremath{\mathbb{R}^{#1}\!}}
\newcommand{\probability}[1]{\ensuremath{\mathbb{P}\left[{#1}\right]\!}}

\newcommand{\thresholdCull}[1]{\ensuremath{\lambda_{#1}}}

\newcommand{\ermdistribution}{\ensuremath{D\!}}
\newcommand{\ermlossSet}[2]{\ensuremath{Q({#1},{#2})}}
\newcommand{\erminstance}[1]{\ensuremath{z_{#1}}}
\newcommand{\ermdatastream}[1]{\ensuremath{Z_{#1}\!}}
\newcommand{\ermparams}[2]{\ensuremath{\vv{w}_{#1}^{#2}}}
\newcommand{\ermparamsSet}{\ensuremath{\mathcal{W}}}

\newcommand{\expected}[2]{\ensuremath{\mathbb{E}_{#1}\left[{#2}\right]}}
\newcommand{\risk}[2]{\ensuremath{\mathcal{R}_{#1}\!\left({#2}\right)}}

\newcommand{\metainstance}[1]{\ensuremath{x_{#1}^{*}}}
\newcommand{\distanceMatrix}[1]{\ensuremath{\text{Dist}_{#1}}}

\newcommand{\bias}{\ensuremath{\mathit{bias}}}
\newcommand{\var}{\ensuremath{\mathit{var}}}
\newcommand{\covar}{\ensuremath{\mathit{covar}}}
\newcommand{\dssubset}[1]{\ensuremath{{X}_{#1}}}
\newcommand{\transpose}{\ensuremath{^{\intercal}}}

\newcommand{\affinity}[1]{\ensuremath{\Delta_{#1}}}

\newcommand{\pai}[1]{\ensuremath{\theta_{#1}}}
\newcommand{\pavec}[2]{\ensuremath{\overrightarrow{\Theta}({#1},{#2})}}
\newcommand{\singular}{\ensuremath{\Sigma}}

\newcommand{\SVD}[1]{\ensuremath{\text{SVD}({#1})}}
\newcommand{\dist}[2]{\ensuremath{d({#1},{#2})}}
\newcommand{\scaledist}[1]{\ensuremath{\tilde{d_{#1}}}}

\newcommand{\metalearner}{\ensuremath{F}}
\newcommand{\reducedpcs}[1]{\ensuremath{{\tilde{U}}_{#1}}}

\providecommand{\keywords}[1]
{
  \small	
  \textbf{\textit{Keywords---}} #1
}
\title{Conceptually Diverse Base Model Selection for Meta-Learners in Concept Drifting Data Streams}

\date{}

\author{Helen McKay\\ 
    Department of Computer Science\\
    University of Warwick\\
    Coventry, UK\\
    \texttt{H.McKay@warwick.ac.uk} \\
    \and
    Nathan Griffiths\\
    Department of Computer Science\\
    University of Warwick\\
    Coventry, UK\\
    \and
    Phillip Taylor\\
    Department of Computer Science\\
    University of Warwick\\
    Coventry, UK\\
}

\begin{document}
\maketitle

\begin{abstract}
    Meta-learners and ensembles aim to combine a set of relevant yet diverse base models to improve predictive
    performance. However, determining an appropriate set of base models is challenging,
especially in online environments where the underlying distribution of data can change over time. In this paper, we
present a novel approach for estimating the conceptual similarity of base models, which is calculated using the Principal
Angles (PAs) between their underlying subspaces. We propose two methods that use conceptual similarity as a metric to
obtain a relevant yet diverse subset of base models: (i) parameterised threshold
culling and (ii) parameterless conceptual clustering. We evaluate these methods against thresholding using common
ensemble pruning metrics, namely predictive performance
and Mutual
Information (MI), in the context of online Transfer Learning (TL), using both synthetic and real-world data.
Our results show that conceptual similarity thresholding has a reduced computational overhead, and yet yields comparable
predictive performance to thresholding using
predictive performance and MI. Furthermore, conceptual clustering achieves
similar predictive performances without requiring parameterisation, and achieves this with lower computational
overhead than thresholding using predictive performance and MI when the number of base models becomes large.
\end{abstract}


\section{Introduction}
Learning in online data streams can be challenging as data availability may be limited if a rich history of observations
cannot be retained~\cite{Tsymbal2004}. Additionally, in dynamic, non-stationary environments, the underlying
distribution of data,
and the mapping from observations to response variables, referred to as a concept, can change over time~\cite{Gama2014}, a phenomenon known as concept
drift~\cite{Zliobaite2010}. 
Meta-learners and ensembles can be used to improve predictive performance in online environments by
combining models learnt historically throughout the data stream~\cite{Bifet2007}. Historical models, which we refer to
as base models, can
be learnt to represent each concept encountered as the data stream progresses. Base models can then be used as input to a meta-learner or
ensemble to enhance predictive capabilities~\cite{Mckay2019}. However, as the data stream progresses, and more base models are
learnt, meta-learners and ensembles can become prone to overfitting, particularly when data availability is limited and
the number of base models is large~\cite{Mckay2020}. 

To improve generalisation, a relevant yet diverse subset of base models can
be selected to be used as input to the meta-learner or ensemble~\cite{Brown2005}. Base model selection is
often achieved within offline environments using metrics such as predictive performance to indicate relevancy, and
Mutual Information (MI) between base model predictions to indicate pairwise diversity~\cite{Dutta2009}. However, determining relevancy and
diversity can be more challenging in online environments due to concept drift~\cite{Brzezinski2016}. 
Encountering concept drift can impact the relevancy of a base model, and therefore requires the relevancy of base models
to be continually re-evaluated as the data stream progresses. Many diversity metrics, such as MI, must also be
recalculated due
to their dependency on the current underlying distribution of data when identifying the covariance between model
predictions. Repeatedly recalculating relevancy and diversity metrics may be computationally
undesirable when using meta-learners or ensembles in data streams, particularly when drifts are encountered frequently or a
large number of potential base models are available.

This challenge is not limited to learning in single online data streams. Meta-learners and ensembles can also be using
in online Transfer Learning (TL)~\cite{Zhuang2021}. Online TL allows base models to be learnt from different sources of data and
combined through the use of a meta-learner or ensemble to improve predictive performance~\cite{Zhao2014}. The
Bi-directional Online Transfer Learning (BOTL) framework allows
knowledge transfer to be conducted bi-directionally across multiple online data streams~\cite{Mckay2019,Mckay2020}. Base
models are created to represent the concepts encountered within a data stream, and are transferred so that they can be
used to improve the predictive performance in other data streams~\cite{Mckay2019}.
The number of base models transferred can quickly become large when the number of data streams in the framework is
large, or when
concept drifts are frequent. Therefore, selecting a relevant yet diverse subset of base models is
required to prevent overfitting~\cite{Mckay2020}.

The contributions of this paper are:
\begin{enumerate}[(i)]
    \item a novel approach for estimating the conceptual similarity of base models in concept drifting data streams,
    \item the parameterised thresholding method for model culling, and the parameterless conceptual clustering method
        for obtaining a subset of relevant yet diverse base models using conceptual similarity, and
\item the use of parameterless conceptual clustering to reduce the number of models transferred in the BOTL
    framework.
\end{enumerate}

Our conceptual similarity metric is determined using the Principal Angles (PAs) between the subspaces in which each base
model was created. Thus, the similarity between pairs of base models remains static in the presence of concept drift and
does not require recalculation. We present empirical results applying our methods in combination with BOTL using two synthetic data generators, namely a
drifting hyperplane emulator and a smart home heating simulator, and real-world data predicting Time-To-Collision (TTC)
from vehicle telemetry. Base models are obtained using three underlying
concept drift detection strategies, specifically Reactive Proactive drift detection (RePro)~\cite{Yang2005}, Adaptive
Windowing (ADWIN)~\cite{Bifet2007}, and Adaptive Windowing with Proactive drift detection (AWPro)~\cite{Mckay2020}.
BOTL~\cite{Mckay2020} is used to transfer models bi-directionally between online data streams. 
We use the thresholding and clustering approaches proposed in this paper to select a
subset of base models from those learnt locally within a data stream, and received from other data streams for
the Ordinary Least Squares (OLS) meta-learner in BOTL.

We show that conceptual similarity thresholding obtains predictive performances comparable to the existing approaches
of performance and MI thresholding, while reducing the computational overhead associated with base models comparisons.
Additionally, conceptual clustering
achieves similar predictive performances without the need for a user defined culling parameter. Although conceptual
clustering
increases the computational overhead in comparison to conceptual similarity thresholding, it can be less
computationally expensive than thresholding using diversity metrics that are not static, such as MI, when the number of base models is
large. We also show
that conceptual clustering can be used to reduce the number of models transferred in online TL.

The remainder of this paper is organised as follows. Section~\ref{relatedwork} outlines existing research relating to
the creation of base models, combining base models, and selecting base models in online environments. Section~\ref{conceptSimilarity}
defines our novel approach of estimating the conceptual similarity between base models. In Section~\ref{baseModelSelect}
we present our methods for selecting a subset of relevant yet diverse base models to be used as input to a
meta-learner, namely (i) parameterised thresholding and (ii) parameterless conceptual clustering. This section
also discusses how
conceptual similarity clustering can be used to reduce the number of models transferred between online data streams in
an online TL framework. 
In Section~\ref{section:motivation} we
provide more insight into why the subset of base models used by a meta-learner should be relevant yet diverse.
Section~\ref{setup} provides details about the BOTL framework used to evaluate our 
techniques, and discusses the datasets used for empirical evaluation. 
Section~\ref{results} presents our empirical results of using parameterised thresholding and parameterless conceptual
clustering as methods to select base models for the OLS-meta-learner in the BOTL framework. Results are also presented
that show conceptual clustering can be used to reduce the number of models
transferred between data streams in online TL. Finally, Section~\ref{conclusion} concludes the paper,
outlining our key findings and the possibilities for future work.

\section{Related Work}\label{relatedwork}

Three important issues must be addressed when using meta-learners or ensembles in online environments. Firstly, base
models must initially be created in online data streams. Secondly, base models must be combined to obtain an overarching
prediction. Thirdly, if the number of base models becomes large in comparison to the available data, a subset of base
models must be selected to be used by the meta-learner or ensemble to prevent overfitting.

\subsection{Creating base models}
When learning in online environments, predictive models must be updated or relearnt in order to adapt to concept
drift~\cite{Tsymbal2004}. Two common approaches to handling concept drift include incremental learners~\cite{Gama2014},
and Concept Drift Detection strategies (CDDs)~\cite{Tsymbal2004}.

Incremental learners gradually update models parameters as new instances are observed in the data
stream~\cite{Tsymbal2004}. This
allows incremental learners to react to concept drift. However, model parameters can be influenced by historical
instances associated with a previously encountered concept~\cite{Koychev2000}. Instance weighting and forgetting mechanisms can be
used to prevent historical instances from negatively impacting predictive performance when concept drifts are
encountered~\cite{Gama2014}. These techniques enable recently observed instances to have a greater influence on model parameters
compared to historical instances, which may not be associated with the current concept~\cite{Jaber2013}. Although incremental
learners can be used in online environments, the effect of continually updating model parameters, without
the ability to detect occurrences of concept drift,
means that knowledge of previously encountered concepts can be lost as the data stream progresses~\cite{Lu2019}. This can be
detrimental in environments that contain recurring concepts since the model parameters for a recurring concept must
be re-learnt~\cite{Yang2006}.

Alternatively, CDDs can be used in online learning environments~\cite{Bifet2007,Mckay2019,Yang2005}. CDDs typically learn a predictive model
over a small window of recent observations. Model predictions are monitored over a sliding window of instances, with new
instances added as they are observed~\cite{Tsymbal2004}. Concept drifts are commonly detected by
monitoring the predictive performance~\cite{Yang2005}, or identifying changes to the distribution of predictive
error~\cite{Bifet2007}. When concept drifts are detected, alternative predictive models can be learnt or
reused~\cite{Mckay2020,Yang2006}. 

CDDs typically only retain recent
observations. This means that they are not influenced by historical instances that belong to a previously encountered
concept when detecting drifts or learning model parameters~\cite{Bifet2007}. Since model parameters
are not usually incrementally updated when using CDDs, historical knowledge of previously encountered concepts can be
retained, and in some approaches are reused when recurring concepts are encountered.
For example, RePro~\cite{Yang2005} detects drifts by monitoring predictive
performance. When the performance drops below a threshold, either a new model is learnt from the recent observations
captured in the
sliding window, or a historical model is reused in the presence of recurring concepts~\cite{Ramamurthy2007}.
Other approaches to detecting concept drifts estimate the precise point of drift by monitoring the distribution of
predictive error. ADWIN~\cite{Bifet2007} and AWPro~\cite{Mckay2020} use this approach so that instances belonging
to the previous concept can be discarded prior to learning a new model. ADWIN does not make use of historical models when
encountering recurring concepts, while AWPro prioritises their reuse over learning new
models~\cite{Bifet2007,Mckay2020}.

To learn a model that generalises well when using a CDD, sufficient instances belonging to the new concept must be observed.
During this period, ineffective predictions may be made as the previously learnt model continues to be used. This is
known as the cold start problem, and
introduces a trade-off between reacting quickly to drifts to prevent prolonged use of an ineffective model,
and waiting for sufficient observations in order to learn a model that generalises well~\cite{BifetThesis}. This means
that
predictive models may have to be learnt from small amounts of data. However, in many real-world data streams,
historically encountered concepts may be similar to the newly encountered concept. To improve predictive performance,
the models learnt to represent each of the concepts encountered throughout the data stream
can be used as base models for a meta-learner or ensemble~\cite{Bifet2009,Du2019,Gomes2017,Kolter2007}. This allows previously learnt models to be
used to enhance predictive performance and reduce the impact of the cold start problem, particularly when concepts
re-occur throughout the data stream~\cite{GomesSoares2015,Kolter2005}.

Some real-world applications that require predictions to be made in online environments consist of multiple devices, each
learning similar tasks from independent data streams. 
For example, a low cost sensor may be used to make predictions via
environmental monitoring in one location, while other low cost sensors conduct the same predictive task in different
locations. 
Although each device learns from independent data streams, knowledge may have
been learnt from one data stream that is beneficial to another. Sharing knowledge may improve predictive
performance, and can be achieved through online TL techniques such as BOTL~\cite{Mckay2020}. BOTL uses a CDD to create
base models that represent the concepts encountered in an online data stream. Models are transferred to other data streams
and used alongside locally learnt models as base models. Base model predictions are combined using an OLS meta-learner to
improve predictive performance for unseen instances~\cite{Mckay2019,Mckay2020}. Knowledge transfer is
bi-directional such that knowledge learnt in one
data stream is made available to others, and visa versa~\cite{Mckay2020}. Other online TL frameworks learn base models
in offline environments and transfer them to an online data
stream~\cite{Ge2013,Grubinger2016GOTL,Wu2017,Zhao2014,Zhuang2021}. However, regardless of the environment in which base
models are learnt, meta-learners and ensembles can be used to combine model
predictions to improve the predictive performance~\cite{Brown2005,Du2019,Dutta2009,GomesSoares2015,Wang2003}. In this
paper, we use RePro~\cite{Yang2005},
ADWIN~\cite{Bifet2007}, and AWPro~\cite{Mckay2020} as CDDs to create base models in the BOTL
framework~\cite{Mckay2019,Mckay2020}

\subsection{Combining base models}
In offline environments, the simplest approach to combining base models is to use ensembles with majority rules
voting mechanisms for classification
tasks, or prediction averaging for regression tasks~\cite{Du2019}. However, these approaches give equal weightings to the
predictions of each base model, which may not be appropriate in the presence of concept drifts. For example, when a
drift is encountered, some base models may not make effective predictions for the new concept. Therefore, equally
weighting models in the ensemble may hinder the overall predictive performance. To overcome this, weighted voting
mechanisms can be used~\cite{GomesSoares2015,Kolter2005,Kolter2007}. Weights can be
determined by measuring the predictive performance of each base model over a recent window of
instances in the data stream.
However, weighted voting mechanisms are most beneficial when base models have been learnt to perform the same tasks,
and have comparable success~\cite{Rokach2010}. This may not be the case when learning in online data streams, since models
may have been learnt to represent different concepts and may not perform comparably. Additionally, in a regression
setting, weighted voting mechanisms rely upon the assumption that the response variable to be predicted will remain
within a consistent range across all learning tasks. This may not be the case in online data streams that encounter
concept drift, or when transferring knowledge between data streams in online TL frameworks.

Alternatively, since future concepts are unknown, a meta-learner can be used to
combine the available base models. 
In comparison to weighting mechanisms, the use of a meta-learner typically allows for
better generalisation by correcting for base model biases when some base models consistently perform poorly for
segments of the data stream~\cite{Rokach2010}. Correcting for such bias can greatly impact the overarching predictive
performance since the base models obtained from historical concepts, or transferred from
other data streams via online TL, are not guaranteed to be relevant or
useful when making predictions for the current concept.
As the
data stream progresses, the number of available base models may
increase, and concept drifts may be encountered. Therefore, the weights associated with each base model must be
updated~\cite{Bifet2009,Gomes2017,Grubinger2016GOTL}. 

Increasing the number of base models used by the meta-learner increases its representational capacity. This allows a better
approximation of the underlying distribution of observable data to be made~\cite{Vapnik1995}, which decreases the
training error of the meta-learner, known as the empirical risk. However, when the number of base models becomes large,
the meta-learner may overfit the window of observable data~\cite{Friedman1997}, which can increase the predictive error
for unseen instances, known as the true risk~\cite{Zhou2012}. This often occurs when the representational capacity of
the meta-learner is too
large to be able to effectively learn the meta-learner model parameters from the fixed sized window of observable
data~\cite{Cherkassky2007,Niyogi1994,Vapnik1995}.
In such cases, the empirical risk can be a poor approximation of the true risk~\cite{Cherkassky2007}, which may lead
to poor predictive performance for unseen instances. Therefore, to prevent overfitting, the representational capacity of
the meta-learner can be reduced by selecting a subset of base models to be used as input. To ensure that the meta-learner
makes effective predictions and generalises well for
unseen instances, a relevant yet diverse subset of base models should be selected~\cite{Brown2005}. In this
paper, we consider how relevancy and diversity metrics can be used to select a subset of base models to be used as input
the to OLS meta-learner in the BOTL framework. Further discussion of how the empirical
risk of a meta-learner can be minimised, while remaining a good approximation of the true risk, through the use of
relevancy and diversity, is presented in Section~\ref{section:motivation}.


\subsection{Selecting base models}
Identifying the best subset of base models is challenging in
online environments since future concepts are
unknown, a rich history of data often cannot be retained, and due to the presence of concept drift.
Existing online ensemble techniques, such as Accuracy Weighted Ensemble (AWE)~\cite{Wang2003}, Online Weighted Ensemble
(OWE)~\cite{GomesSoares2015} and Additive Expert ensemble
(AddExp)~\cite{Kolter2005}, select base models using the recency of a model, the predictive performance, or a
combination of the two, as indicators of relevancy. Using only the recency of a model as an indicator of relevancy may
be undesirable in online data streams since a historical model may be more relevant than a recently learnt model due to
recurring concepts~\cite{GomesSoares2015,Kolter2005}. The diversity among base models in online
environments is
considered less frequently. Five diversity metrics for regression ensembles were presented by Dutta,
namely the correlation coefficient between base model predictions, the covariance between model predictions, the
pairwise Chi-square of model predictions, the standard deviation of 
predictions as a disagreement bound, and the MI between model predictions~\cite{Dutta2009}. MI is a
frequently used regression ensemble diversity metric, where the pairwise diversity of base models is measured
using the MI
between predictions of base models over recent observations~\cite{Gomes2017,Mckay2020}. However, each of the measures
of diversity proposed by Dutta~\cite{Dutta2009} are dependent on the recent window of observations used to evaluate
base models. This is problematic in concept drifting data streams and online TL environments since
the diversity of base models that obtain similar predictive performance for some windows of observations, but
have been learnt from different distributions, or learnt to represent different concepts, is not accounted
for~\cite{Gomes2017}. Additionally, due to the dependency on a window of recent observations, measuring diversity
by the level of disagreement
between base model predictions can be influenced by concept drift, and therefore must be recalculated as the data stream
progresses.

To overcome this, we introduce a novel method for measuring the diversity of base models for meta-learners in online
environments, which
allows the pairwise conceptual similarity
between base models to be estimated independently of the current distribution of observable data. This approach uses the
similarity between the underlying subspaces in which each base model was learnt to obtain a measure of diversity that
remains static as the data stream progresses, even in the presence of concept drift.

\begin{table}[ht]
    \caption{Notation}
    \label{notation}
    \centering
    \begin{tabular}{ll}
        \toprule
         & Definition \\
         \midrule
        $\domain{\alpha}$ & Domain $\alpha$\\
        $\datastream{\alpha}$ & Data stream in domain $\alpha$, where $\datastream{\alpha}=\{\instance{1},\dots,\instance{t},\dots,\instance{n}\}$\\ 
        $\instance{t} \in \datastream{\alpha}$ & The $t^{th}$ observed instance in $\datastream{\alpha}$\\
        $\labelspace{\alpha}$ & Response variable space in domain $\alpha$\\
        $\tLabel{t}\in\labelspace{\alpha}$ & The $t^{th}$ response variable in $\labelspace{\alpha}$\\
        $\conceptSet{\alpha}$ & The set of concepts encountered in domain $\alpha$\\
        $\concept{\alpha}{i}\in\conceptSet{\alpha}$ & The $i^{th}$ concept encountered in domain $\alpha$\\
        $\datastreamsegment{\alpha}{i}\in\datastream{\alpha}$ & The data stream segment corresponding to concept
        $\concept{\alpha}{i}$ in domain $\alpha$\\\
        $\window$ & Sliding window of $|\window|$ instances, $\window = \{\instance{t-|\window|},\dots,\instance{t}\}$\\
        $\model{\alpha}{i}:\datastreamsegment{\alpha}{i}\rightarrow\labelsegment{\alpha}{i}$ & Model $i$ learnt in domain $\alpha$\\
        $\pcs{i}$ & Principal Components of the window used to learn model $\model{}{i}$\\
        $\modelset$ & Set of stable, locally learnt and transferred models\\
        $\modelsubset$ & A subset of stable, locally learnt and transferred models, $\modelsubset \subseteq\modelset$\\
        $\metamodel{}: \datastream{\alpha}\rightarrow \labelspace{\alpha}$ & Meta-learner in domain $\alpha$\\
        $\metainstance{t}$ & Meta instance of base model predictions for instance $\instance{t}$\\
        $\predict{t}$ & Prediction using $\metamodel{\modelsubset}(\instance{t})$ where $\modelsubset\subseteq\modelset$\\
        \bottomrule
    \end{tabular}
\end{table}
\section{Conceptual Similarity}\label{conceptSimilarity}
In this section, using the notation in Table~\ref{notation}, we present our novel method for estimating the conceptual
similarity between pairs of base models in order to determine the diversity among base models to be used by
meta-learners in online environments.

Given a set, $\modelset$, of $k$ base models and a data stream, $\dssubset{}$, the meta-learner, $\metalearner$,
must learn a predictive function, mapping the predictions of base models, $\model{}{i}(\instance{t})$, for instance
$\instance{t}\in\dssubset{}$ at time $t$ to the response variable $\tLabel{t}$, such that
$\metalearner\left(\metainstance{t}\right)\rightarrow\tLabel{t}$, where $\metainstance{t}=
\langle\model{}{1}(\instance{t}),\dots,\model{}{k}(\instance{t})\rangle$.
Due to the presence of concept drift, the meta-learner must re-weight base model predictions as the data stream
progresses. Weights are learnt using the predictions of base models over a sliding window of recent observations,
$\window$, which are then used to combine base model predictions for unseen instances. As the data stream progresses,
new models are learnt, and are transferred in the case of online TL, so that the meta-learner can use previously learnt knowledge to aid
predictive performance.

To prevent the meta-learner from overfitting, a relevant yet diverse subset of base models must be selected. Although
the relevancy of a base model is dependent on the current distribution of data observable in $\window$, the pairwise
diversity of
two base models can be considered independently of $\window$. To assess diversity, it may be desirable to measure model
similarity by comparing the underlying distributions of all data belonging to each of the concepts that the base
models were learnt to represent~\cite{Gomes2017}.
However, for many real-world applications, it may not be feasible to retain a rich history of data belonging to each
concept, and the transfer of all data may be impractical due to the communication overhead when used in online
TL. Instead, we approximate conceptual similarity
by considering the subspace similarity using the PAs between the data in which each base model was trained. The PAs
between the subspaces in which a pair of base models were learnt is defined as
follows~\cite{Bjorck1973,Golub1995,Knyazev2012}.

\begin{definition}[Principal Angles (PAs) between subspaces]\label{PAs}
    For two base models $i$ and $j$, let $\dssubset{i}\in\realnumbers{N\times m}$ and
    $\dssubset{j}\in\realnumbers{N\times m}$
    denote the subspaces in which they were learnt, containing $N$ instances, $m-1$ features, and the associated
    response variable, and let
    $\pcs{i}\in\realnumbers{N\times p}$ and $\pcs{j}\in\realnumbers{N\times q}$ represent orthonormal bases of
    $\dssubset{i}$ and $\dssubset{j}$ respectively. Using Singular Value
    Decomposition (SVD) we obtain $\SVD{\pcs{i}\transpose\pcs{j}} = {A}\Sigma{B}$, where ${A}$ and
    ${B}$ are the unitary matrices and $\Sigma\in\realnumbers{p\times q}$ is the diagonal matrix of singular
    values, $\Sigma = \text{diag}([s_{1},\dots,s_{r}])$, where $r=\text{min}(p,q)$~\cite{Knyazev2012}. The PAs
    between subspaces are given by
    \begin{equation}\label{singularangles}
        \pavec{\dssubset{i}}{\dssubset{j}} = [\arccos(s_{1}),\dots,\arccos(s_{r})].
    \end{equation}
\end{definition}

\subsection{Estimating conceptual distance}
The conceptual similarity between the underlying concepts a pair of base models were learnt to represent can be
estimated using PAs. In order to obtain the PAs in Definition~\ref{PAs}, an
orthonormal representation of the subspaces in which each base model was learnt is required. 
The Principal Components (PCs) of a subspace is an orthonormal representation, and can be obtained using Singular Value
Decomposition (SVD) on the covariance matrix, 
\begin{equation}\label{SVDforPCs}
\SVD{\dssubset{i}\transpose\dssubset{i}} = \pcs{i}\singular V,
\end{equation}
such that $\pcs{i}\in\realnumbers{N\times N}$ are the PCs, and $\singular$ their singular values. 

Therefore, to estimate the conceptual similarity of base models, both the model learnt to represent a concept, and an
orthonormal representation of the training data must be stored in online learning frameworks, and transferred
between data streams when using online TL. In these online learning frameworks, the memory and communication
overhead required to store and transfer
orthonormal representations of the subspace associated with each base model, $i$, can be reduced by retaining only the
first $p$ PCs that capture $99.9\%$ of the variance of the original training data, $\dssubset{i}$. The diagonal matrix
of singular values, $\singular$, obtained through SVD in Equation~\ref{SVDforPCs}, can be used to determine the number
of PCs to retain.

Let $\singular=\text{diag}([S_{11},\dots,S_{NN}])$ represent the diagonal matrix of singular values in
Equation~\ref{SVDforPCs}. The number of PCs,
$p$, that capture $99.9\%$ of the variance can be identified using
\begin{equation}\label{selectPCs}
    1 - \frac{\sum_{i=1}^{p}S_{ii}}{\sum_{j=1}^{N}S_{jj}} \leq 0.001.
\end{equation}

The percentage of variance captured by the $p$ PCs can be decreased to reduce the impact of noise in the data stream.
This allows the orthonormal representations, and the PAs between them, to be more robust
to noise. However, alternative methods of
obtaining orthonormal representations of each subspace, such as Laplacian PCA~\cite{Zhao2007} and Robust
PCA~\cite{Xu2010}, can be used to improve robustness to noise and outliers in noisy data streams~\cite{John2019}.
Once $p$ has been identified we can reduce the matrix of $N$ PCs, $\pcs{i}\!\in\!\realnumbers{N\times N}$, such that
only the first $p$ PCs are retained, $\reducedpcs{i}\!\in\!\realnumbers{N\times p}$. As the reduced orthonormal
representation, $\reducedpcs{i}$, captures $99.9\%$ of variance, for ease of notation throughout the remainder of this
paper we use $\pcs{i}$ to denote the reduced matrix of PCs. Therefore, from this point
forward, $\pcs{i}\!\in\!\realnumbers{N\times p}$. 

Using Definition~\ref{PAs} we can calculate the PAs between the subspaces in which two base models $i$ and $j$ were
learnt, $\dssubset{i}$ and $\dssubset{j}$, using the reduced PCs, $\pcs{i}$ and $\pcs{j}$, as orthonormal bases. We
define the conceptual distance between base models $i$ and $j$ as follows.

\begin{definition}[Conceptual distance between base models]\label{def:PAcalc}
    Let $\pavec{\dssubset{i}}{\dssubset{j}}\in\realnumbers{r}$ be the vector of PAs, obtained using
    Definition~\ref{PAs}, between the PCs
    $\pcs{i}$ and $\pcs{j}$ for $\dssubset{i}$ and $\dssubset{j}$ respectively. The conceptual distance between base
    models $i$ and $j$ is defined as
    \begin{equation}\label{PAdistance}
    \displaystyle
        \dist{i}{j} = \frac{1}{r}\left(\sum_{h=1}^{r}\left(1-\cos(\pai{h})\right)\right) = 1 -
        \frac{1}{r}\left(\sum_{h=1}^{r}\cos(\pai{h})\right),
    \end{equation}
    where $\pai{h}\in\pavec{\dssubset{i}}{\dssubset{j}}$ and $r=\text{min}(p,q)$. Therefore, $\dist{i}{j}\rightarrow0$
    when base models $i$ and $j$ have been learnt from similar subspaces, and $\dist{i}{j}\rightarrow1$ when subspaces
    are dissimilar.
\end{definition}

\subsection{Estimating conceptual similarity}
Finally, to estimate the conceptual similarity among base models, we create an affinity matrix.
\begin{definition}[Conceptual similarity]\label{def:affinityMatrix}
    The affinity matrix, $\affinity{}\in\realnumbers{|\modelset|\times|\modelset|}$, where $|\modelset|$ is the number
    of available base models, is given by
    \begin{equation}\label{affinityMatrixCalc}
        \affinity{ij}=
        \begin{cases}
            \text{exp}(\frac{-{\dist{i}{j}}^{2}}{\scaledist{i}\scaledist{j}}) & \text{if } i\neq j\\
            0 & \text{otherwise,} 
        \end{cases}
    \end{equation}
    where $\scaledist{i}$ and $\scaledist{j}$ are local scaling parameters, which allows the
    conceptual difference between base models to be scaled by the surrounding neighbourhoods of $i$ and
    $j$~\cite{Zelnik2005} such that $\scaledist{i}=\dist{i}{k}$, where $k$ is the $k^{th}$ nearest neighbour of base model $i$.
\end{definition}
Zelnik-Manor and Perona~\cite{Zelnik2005} suggest that a value of $k=7$ yields
    good results for local scaling, even for high-dimensional image segmentation and document classification tasks.
    However, since the dimensionality of the affinity matrix for base models
    is likely to be
    low in comparison to the affinity matrices used by Zelnik-Manor and Perona~\cite{Zelnik2005}, we investigated the
    impact of
    using $k$ values between 2 and 7. The results of using different local scaling parameters are briefly discussed in
    Section~\ref{results}, and we show that these scaling parameter values obtain similar results, therefore in general
    we use $k=7$.
    Local scaling is used to allow better affinities to be obtained when the density of conceptually similar base models
    varies~\cite{Zelnik2005}. Although this requires a parameter, $k$, to be defined, it has been shown that parameter
    tuning is not typically needed for local scaling to perform well~\cite{John2019}. Alternatively, other scaling
    techniques could be
    used, such as density-aware kernels~\cite{John2019,Zhang2011}, to amplify intra-cluster similarities in order to
    account for locally dense areas of conceptually similar base models~\cite{John2019}.

Using Definition~\ref{def:affinityMatrix}, we obtain the affinity matrix, $\affinity{}$, where element
$\affinity{ij}\rightarrow 1$ when base models $i$ and $j$ are conceptually similar, and $\affinity{ij}\rightarrow 0$
when they are dissimilar. 

\section{Base Model Selection}\label{baseModelSelect}
To select a relevant yet diverse subset of base models, we propose (i) parameterised thresholding and (ii)
parameterless conceptual clustering, using the estimated conceptual similarity of base models. We also consider how conceptual
clustering can be used to determine when a newly learnt base model should be transferred to other data streams in online
TL settings.

\subsection{Parameterised thresholding}
\begin{algorithm}[t]
    \DontPrintSemicolon
    \caption{Concept similarity thresholding}
    \label{algo:thresholding}
    \KwInput{$\modelset$, $\model{}{i}$, $\pcs{i}$, $\distanceMatrix{}$, $\thresholdCull{CS}$}
    \For{$j \in \modelset$}{
        $\distanceMatrix{i,j}$ = getDistances($\pcs{i},\modelset.\text{getPCs}(j)$) using Def~\ref{PAs} and
        Def~\ref{def:PAcalc}\;
        $\distanceMatrix{j,i} = \distanceMatrix{i,j}$\;
        
    }
    iAffinities = getAffinities($\distanceMatrix{},i,\modelset$) using Def~\ref{def:affinityMatrix}\;
    \eIf{$\forall j \in \text{iAffinities} < \thresholdCull{CS}$}{
        Add $\{\model{}{i},\pcs{i}\}$ to $\modelset$\;
    }{
        $\forall j\in\modelset:$ Remove $\distanceMatrix{i,j}$ and $\distanceMatrix{j,i}$ from $\distanceMatrix{}$\;
    }
    \KwRet{$\modelset$}
\end{algorithm}
As a data stream progresses, new models are made available when encountering concept drifts, or are received from
other domains via online TL. The pairwise conceptual similarities between a new model, $\model{}{i}$, and existing base
models,
$\model{}{j}\in\modelset$, are calculated using the average PA between the two subspaces in which each model was learnt (Definitions~\ref{PAs}
and~\ref{def:PAcalc}), to obtain a locally scaled affinity metric (Definition~\ref{def:affinityMatrix}). Given a user defined
culling threshold, $\thresholdCull{CS}$, models are added to $\modelset$ if no existing base model in
$\modelset$ is considered conceptually similar. This means that in order for a new model, $\model{}{i}$, to be used as
input to the meta-learner, its affinity to all other available base models, $\forall j \in \modelset:\affinity{ij}$, is
less than the conceptual similarity threshold, $\thresholdCull{CS}$. Since conceptual similarity is calculated
independently of the
underlying distribution of data, a new model, $\model{}{i}$, with an
affinity to an existing base model greater than $\thresholdCull{CS}$ can be discarded and does not need to be
reconsidered as input to the meta-learner. This
process, shown in Algorithm~\ref{algo:thresholding}, obtains a diverse subset of base models. 

Using a static diversity metric such as conceptual similarity prevents the need to re-evaluate base models since they do
not need to be reconsidered once culled. However, the relevancy of remaining
base models is dependent on the current concept. To ensure that a relevant subset of the remaining base models are used
by the meta-learner, the predictive performance of each base model is evaluated to create the subset of base models,
$\modelsubset$, that are used by the meta-learner to make predictions for unseen instances. The predictive performance
of the models that remain after executing Algorithm~\ref{algo:thresholding}, $\modelset$, are evaluated over the current
sliding window of data, $\window$. Those that
achieve an $R^{2}$ performance greater than a performance threshold, $\thresholdCull{\text{perf}}$, are included in
$\modelsubset$. Base models that achieve an $R^{2}$ performance less than $\thresholdCull{\text{perf}}$ are
temporarily excluded from the meta-learner until a concept drift is encountered. This ensures that
the subset of base models used by the meta-learner remains relevant to the current concept.

Selecting culling parameters can be challenging as they are dependent on the presence of noise in the
underlying data stream, the number of possible base models, and their separability for a given similarity
metric~\cite{Mckay2020}. Values that promote aggressive culling may result in discarding base models that are beneficial to
the meta-learner, while values that are not aggressive enough may cause overfitting~\cite{Mckay2020}. 

To overcome this, culling parameters could be updated as the data stream progresses using cross-validation. However,
this would require the meta-learner to be validated for various threshold values over a small window of data.
This increases computation, and the performance obtained by the meta-learner through cross-validation would
likely be an overestimate since validation splits are unlikely to be independent due to the dependencies between
consecutive instances within online data streams~\cite{Hammerla2015}. This means that considerable domain expertise is
required to select the culling parameters for meta-learners in online environments.

\subsection{Parameterless conceptual clustering}\label{paramConceptualClustering}
To prevent the need for domain expertise, we introduce parameterless conceptual clustering. As with conceptual
similarity thresholding, when new models are learnt in a data stream or received via online TL, the affinity between
existing base models, $\model{}{j}\in\modelset$, and the new model, $\model{}{i}$, must be calculated using
Definitions~\ref{PAs},~\ref{def:PAcalc} and~\ref{def:affinityMatrix}. Once the affinity matrix, $\affinity{}$, has
been obtained we can consider it as a fully connected graph, where nodes represent available base models, and edges
represent their pairwise conceptual similarity. This allows graph clustering
algorithms, such as Spectral Clustering (SC), to be used to identify groups of conceptually similar base models.
SC algorithms typically have complexity $O(n^2)$ to create the similarity matrix, and $O(n^3)$ for spectral
analysis~\cite{Yang2019}, where $n$ is the number of available base models, $|\modelset|$. Therefore, it is important to
use a static similarity metric, such as conceptual similarity, so that updating the similarity matrix and clustering
available base models is only required when a new base model is learnt, or received from another data stream. Using
metrics such as MI would require both the similarity matrix and spectral analysis to be repeatedly updated due to the
dependency on the current distribution of observable data. The computational complexity of this makes the use of metrics
such as MI for clustering similar base models infeasible for most real-world applications.

We use Self-Tuning Spectral Clustering (STSC), a well known SC algorithm~\cite{Zelnik2005}, which allows the number of
clusters to be determined automatically. STSC uses the local scaling
in Equation~\ref{affinityMatrixCalc}, and incrementally rotates the eigenvectors traditionally obtained from SC to
estimate the number of clusters~\cite{Zelnik2005}. Automatically determining the number of clusters is advantageous when
considering the diversity among base models in online environments, particularly in online TL, where it is not known
how many concepts will be encountered in each data stream, or how similar the concepts learnt from different data streams will be.

\begin{algorithm}[t]
    \DontPrintSemicolon
    \caption{Conceptual clustering base model selection}
    \label{algo:conceptclust}
    \KwInput{$\window$, $\modelset$, $\model{}{i}$, $\pcs{i}$, $\distanceMatrix{}$, $\affinity{}$}
    \For{$j \in \modelset$}{
        $\distanceMatrix{i,j}$ = getDistances($\pcs{i},\modelset.\text{getPCs}(j)$) using Def~\ref{PAs} and
        Def~\ref{def:PAcalc}\;
        $\distanceMatrix{j,i} = \distanceMatrix{i,j}$\;
    }
    $\affinity{}$=getAffinityMatrix($\distanceMatrix{}$) using Def~\ref{def:affinityMatrix}\;
    clusterGroups = STSC($\affinity{}$)~\cite{Zelnik2005}\;
    \For{$c \in \text{clusterGroups}$}{
        Add bestPerformingModel($c,\window$) to $\modelsubset$\;
        \If{bestPerformingModel($c,\window$) is $\model{}{i}$}{
            Also add secondBestPerformingModel($c,\window$) to $\modelsubset$\;
        }
    }
    \KwRet{$\modelsubset$}
\end{algorithm}

Once clusters of conceptually similar base models have been identified using Algorithm~\ref{algo:conceptclust}, we
create the subset of relevant yet diverse base models, $\modelsubset$, to be used as input to the meta-learner by
selecting 
one base model from each cluster. To select a base model from each cluster, we
evaluate their predictive performance on the current window of observable data, $\window$, and select the model with the
highest $R^{2}$ performance as an indicator of relevancy. In the case where the best performing model in a cluster is
the local model, $\model{}{i}$, that has been learnt to represent the current concept, we also add the second best
performing model in that cluster to $\modelsubset$. This ensures that the meta-learner can benefit from the additional
support of a model learnt historically, or from another data stream, that is conceptually similar to the current
concept. 

\subsection{Reducing Transfer in Online TL}
Parameterless conceptual clustering can also be used to reduce the number of models transferred in online TL frameworks
such as BOTL. We
achieve this by clustering locally learnt models prior to knowledge transfer.
When a new model is learnt locally, it is only
transferred to other data streams if it is assigned to a cluster that does not contain a model that has previously been
transferred. Although this increases computation prior to transfer\footnote{SC has complexity $O(n^3)$, where $n$ is the
number of base models.},
it reduces the communication overhead by preventing
two conceptually similar models, learnt from the same data stream, from being transferred. This also reduces computation
in the data streams receiving
transferred models by minimising the number of conceptually similar base models that must be evaluated in order
to select a subset of base models to be used as input to the meta-learner.

\section{Minimising Meta-Learner Risk using Relevancy and Diversity}\label{section:motivation}
Before presenting experimental results for our base model selection strategies, we consider how the empirical risk and
true risk of a meta-learner is impacted by the number of available base models, and the importance of selecting
a subset of relevant yet diverse base models as input to a meta-learner in an online environment.

\subsection{Increasing the Number of Base Models}
In this section we consider the effect of increasing the number of base models available to a meta-learner. We use the
meta-learner in the BOTL framework to demonstrate this, where base models are learnt from each data stream using an
underlying CDD. Further details on the BOTL framework and underlying CDDs are presented in Section~\ref{setup}.

Increasing the number of base models used by the meta-learner reduces empirical risk. This can be seen by considering
the squared loss when using a single model, $\model{}{i}$, that has been locally learnt by the underlying CDD to
represent the concept currently observed in the data stream without influence from historically learnt models, or models
transferred from other data streams. This is equivalent to
constraining the OLS meta-learner optimisation problem such that all base model weights are $0$, except for
$\model{}{i}$, which is given weight $1$, such that
\begin{equation}\label{OLSpred}
    \metamodel{\modelset}\left(\metainstance{t}\right) = w_{0} +
    \sum_{j=1}^{j=k-1}w_{j}\model{}{j}\left(\instance{t}\right) + w_{i}\model{}{i}\left(\instance{t}\right)\text{, where
        $j\neq i$},
\end{equation}
and
\begin{equation}\label{constrainedOpt}
{\metamodel{\modelset}}^{*}\left(\metainstance{t}\right) =
\left(0+\displaystyle\sum_{j=1}^{j=k-1}0\model{}{j}\left(\instance{t}\right)
    +1\model{}{i}\left(\instance{t}\right)\right),
\end{equation}
where $\metamodel{\modelset}$ is the unconstrained meta-learner, which learns weights $w_{0},\dots,w_{k}$ for each
of the $k$ base models, and ${\metamodel{\modelset}}^{*}$ is the constrained meta-learner, where
only the current model from the underlying CDD is used to make predictions~\cite{Mckay2020}. To obtain the weights assigned to the $k$
base models in
Equation~\ref{OLSpred}, the OLS meta-learner must solve the optimisation problem,
\begin{equation}\label{OLSopt}
\underset{w_{0},\dots,w_{k}}{\text{min}}\sum_{t=1}^{|\window|} \left(\tLabel{t} -
        \left(w_{0}+\sum_{j=1}^{j=k-1}w_{j}\model{}{j}\left(\instance{t}\right)+w_{k}\model{}{i}\left(\instance{t}\right)\right)\right)^2.
\end{equation}
Since the optimisation problem in Equation~\ref{OLSopt} is convex, the empirical risk of the unconstrained meta-learner,
$\metamodel{\modelset}$, which uses the predictions of all base models as inputs, is less than the empirical risk of the
constrained meta-learner, ${\metamodel{\modelset}}^{*}$, which uses the locally learnt model alone. This occurs because
the representational capacity of the meta-learner increases, allowing more complex underlying distributions in the data
stream to be captured.

\begin{figure}[tb]
\centering
\begin{subfigure}[b]{0.49\textwidth}
    \centering
\centerline{\includegraphics[width=\columnwidth]{{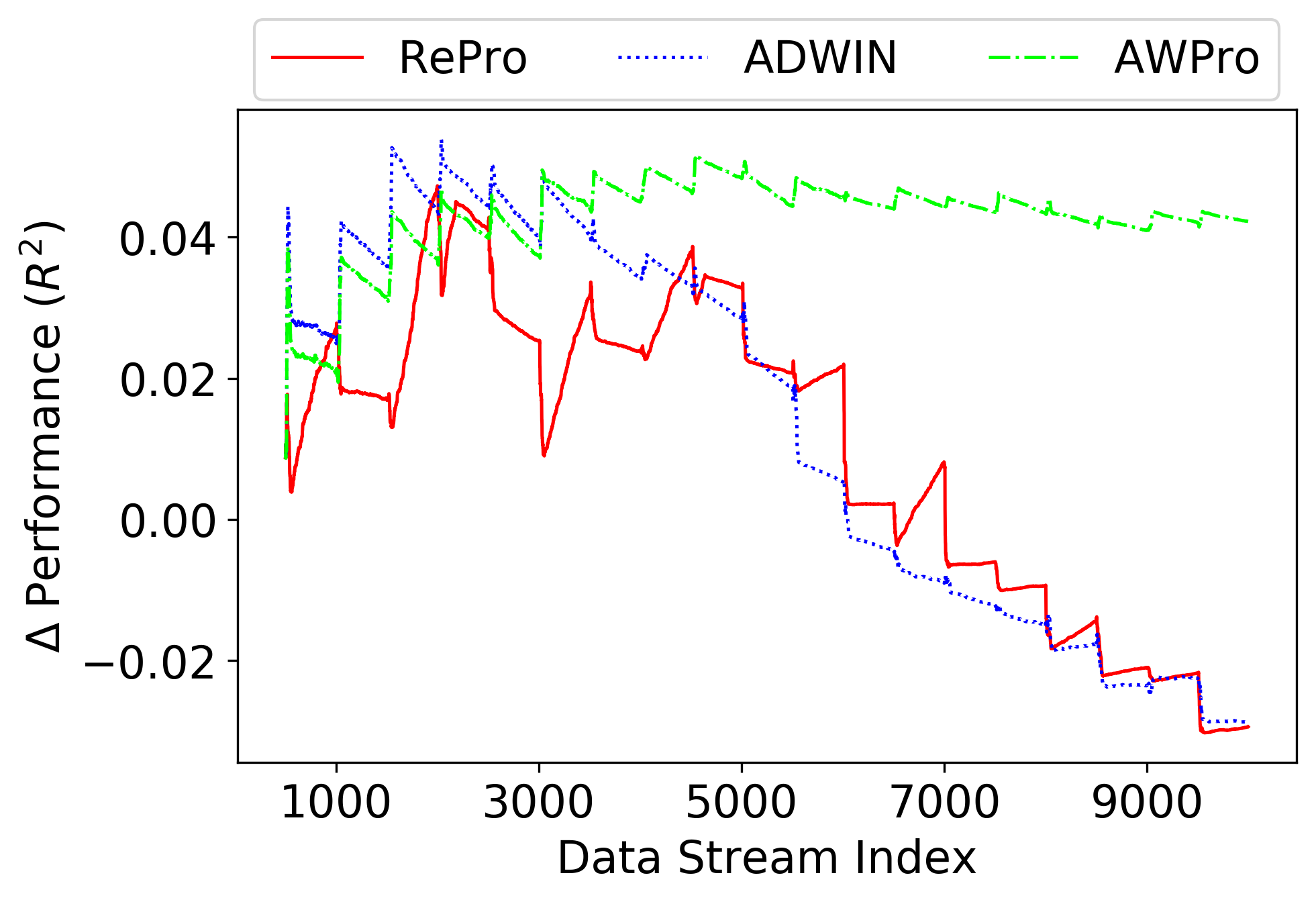}}}
\caption{Difference in $R^2$ predictive performance.}
\end{subfigure}
\hfill
\begin{subfigure}[b]{0.49\textwidth}
    \centering
\centerline{\includegraphics[width=\columnwidth]{{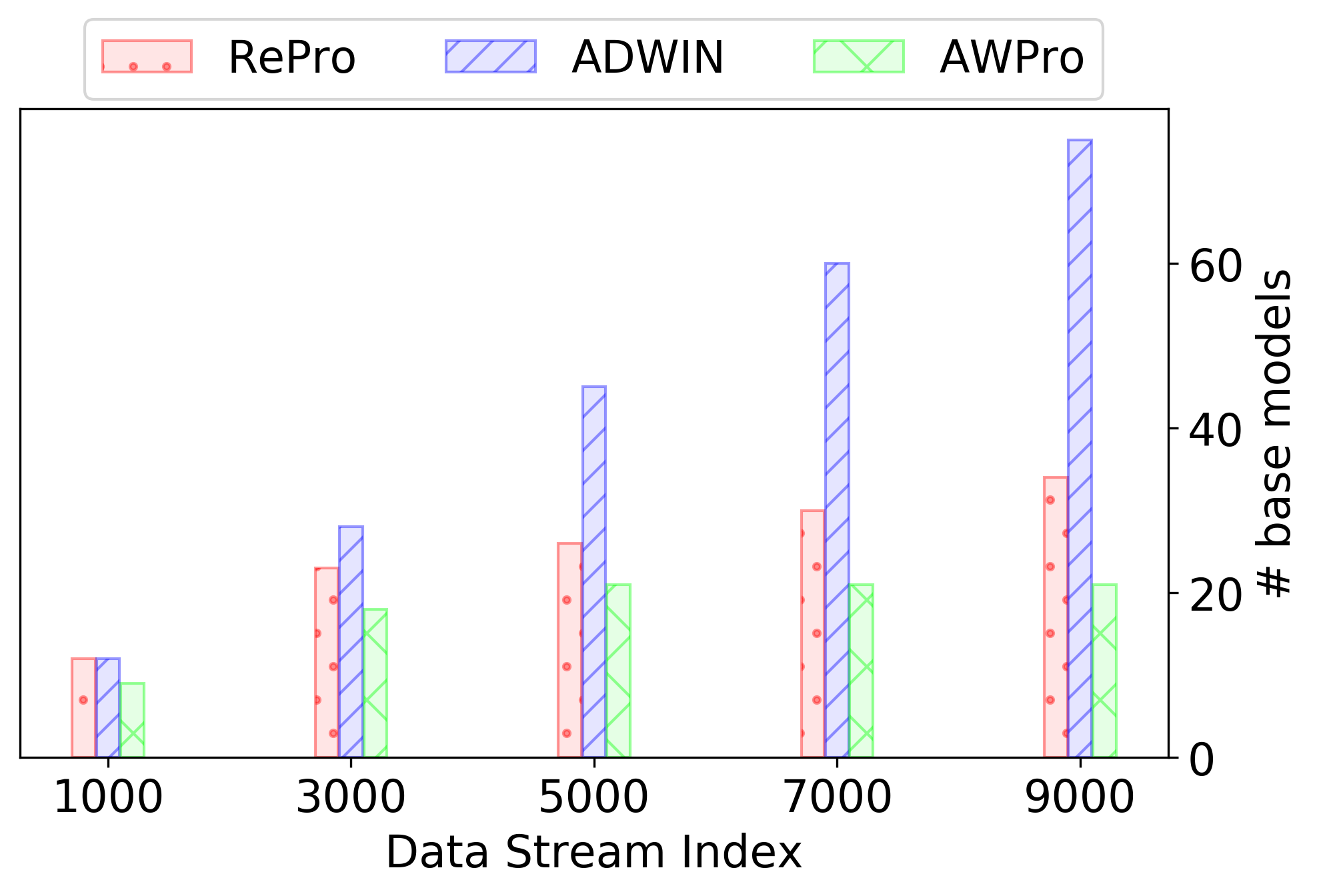}}}
\caption{Number of base models.}
\end{subfigure}
\caption{BOTL vs. CDDs: The difference in $R^2$ performance (a) between the OLS meta-learner in BOTL vs. the underlying
    CDDs of RePro, ADWIN, and AWPro, and the number of models used as base models (b) for a drifting hyperplane data
    stream with sudden drifts. Base models are learnt locally and transferred from 4 other sudden drifting hyperplane
    data streams.}
\label{botlperf}
\end{figure}

As the data stream progresses, the number of available base models increases as new models are learnt locally using the CDDs,
or transferred via online TL. Since the window of available data remains fixed in size, the likelihood of 
overfitting increases as the number of base models becomes large in comparison to the window of available
data~\cite{Friedman1997}. This is illustrated in Figure~\ref{botlperf}, which shows the difference in predictive
performance of the OLS meta-learner used by BOTL compared to using the model learnt via the underlying CDD alone in a
sudden drifting data stream\footnote{A description of the sudden drifting hyperplane dataset is provided in
    Section~\ref{setup}.}. Combining base model predictions initially improves predictive performance. However, as
the number of base models grows, the meta-learner becomes more susceptible to overfitting, causing a
reduction in predictive performance, eventually resulting in worse performance than using the current locally learnt
model alone. However, when using AWPro as the underlying CDD, we do not observe a decrease in performance as the data
stream progresses. This is because AWPro prioritises the reuse of historical models in the presence of recurring
concepts, reducing the number of base models available to the meta-learner in comparison to ADWIN~\cite{Mckay2020}.
AWPro also estimates the precise point of drift prior to learning a new model. This prevents instances belonging to a
previous concept impacting the
detection of recurring concepts, reducing the number of base models available to the meta-learner
in comparison to RePro~\cite{Mckay2020}.
\begin{table}[tb]
    \caption{ERM Notation}
    \label{ermnotation}
    \centering
    \begin{tabular}{ll}
        \toprule
         & Definition \\
         \midrule
         $\risk{}{\ermparams{}{}}$ & True risk of model with parameters $\ermparams{}{}$\\
         $\erminstance{t} = (\instance{t},\tLabel{t})$& Instance $\instance{t}$ at time $t$ and respective response
         variable $\tLabel{t}$\\
         $\ermlossSet{\erminstance{t}}{\ermparams{}{}}$& Loss function of a model parameterised by $\ermparams{}{}$ for
         instance $\erminstance{t}$\\
         $\ermparams{0}{}$& Optimal parameters for a given function\\
         $\ermdatastream{|\window|}$&Training sample of size $|\window|$\\
         $\risk{\text{emp}}{\ermparams{|\window|}{}}$&Empirical risk of model with parameters
         $\ermparams{|\window|}{}$ learnt over $\ermdatastream{|\window|}$\\
         $\ermparams{|\window|}{\ast}$&Model parameters that minimise the empirical risk over
         $\ermdatastream{|\window|}$\\
        \bottomrule
     \end{tabular}
 \end{table}

Using the principal of Empirical Risk Minimisation (ERM)~\cite{Cherkassky2007}, and the notation in
Table~\ref{ermnotation}, we can consider factors that impact the
meta-learner's ability to generalise well for unseen instances in a data stream between concept drifts.

\subsection{ERM for Meta-Learners in Online Environments}

In ERM, the overarching aim is to learn a function that maps the input space,
$\datastream{}$, to the response variable space, $\labelspace{}$, drawn from some unknown distribution. If the
distribution was known, we could find the function parameters, $\ermparams{}{}$, that minimise the risk,
$\risk{}{\ermparams{}{}}$, such that
\begin{equation}
    \begin{aligned}
    \risk{}{\ermparams{}{}}& = \int\ermlossSet{\erminstance{}}{\ermparams{}{}}d\ermdistribution\\
    & = \sum\ermlossSet{\erminstance{}}{\ermparams{}{}}\\
    \risk{}{\ermparams{0}{}}& = \min_{\ermparams{}{}\in\ermparamsSet}\risk{}{\ermparams{}{}},
\end{aligned}
\end{equation}
where $\erminstance{}$ is an instance and response variable pair, $\erminstance{} = (\instance{},\tLabel{})$, and
$\ermlossSet{\erminstance{}}{\ermparams{}{}}$ is the loss function, parameterised by $\ermparams{}{}$, on instance
$\erminstance{}$. For example, the squared loss of instance $\instance{t}$ is
$\ermlossSet{\erminstance{}}{\ermparams{}{}} = (\tLabel{t} - \model{}{}(\instance{t}))^{2}$ where model $\model{}{}$ has
parameters $\ermparams{}{}$. Therefore, $\risk{}{\ermparams{}{}}$ is the risk, or loss, of a model parameterised by
$\ermparams{}{}$. We use $\risk{}{\ermparams{0}{}}$ to denote the risk of a model with optimal parameters,
$\ermparams{0}{}$, which minimise the risk
over the known distribution, $\ermdistribution$. However, since the distribution is unknown,
ERM approximates the optimal risk by evaluating the loss over a window, $\window$, of training samples,
$\ermdatastream{|\window|}$, to obtain the
empirical risk, $\risk{\text{emp}}{\ermparams{|\window|}{\ast}}$. The empirical risk is defined as,
\begin{equation}
    \risk{\text{emp}}{\ermparams{|\window|}{\ast}} = \sum_{i=1}^{|\window|}\ermlossSet{\erminstance{i}}{\ermparams{|\window|}{\ast}}.
\end{equation}

The empirical risk is inherently biased towards the training sample, $\ermdatastream{|\window|}$, and therefore the
function learnt often underestimates
the risk of the same function when used on unseen instances of data belonging to the same concept. The risk over unseen
instances is known as the
true risk, and is denoted by $\risk{}{\ermparams{|\window|}{\ast}}$~\cite{Cherkassky2007}. This means that for
a given training sample
\begin{displaymath}
    \risk{\text{emp}}{\ermparams{|\window|}{\ast}} < \risk{}{\ermparams{|\window|}{\ast}}.
\end{displaymath}

Due to the randomness of $\ermdatastream{|\window|}$, the empirical risk and true risk ,$\risk{\text{emp}}{\ermparams{|\window|}{\ast}}$ and
$\risk{}{\ermparams{|\window|}{\ast}}$, are also random sequences, and the law of large numbers states that the average
converges to its expected value as the number of samples grows large~\cite{Cherkassky2007,Niyogi1994,Vapnik1995}.
Therefore, as
$|\window|\rightarrow\infty$,
\begin{equation}\label{limits}
    \begin{aligned}
    \risk{}{\ermparams{|\window|}{\ast}} &\rightarrow \risk{}{\ermparams{0}{}},\\
    \risk{\text{emp}}{\ermparams{|\window|}{\ast}} &\rightarrow \risk{}{\ermparams{0}{}}.
\end{aligned}
\end{equation}
As the number of training samples grows, a better estimate of the parameter values $\ermparams{|\window|}{\ast}$
can be obtained for a fixed number of parameters $k$, where $\text{\textbar}\ermparams{|\window|}{\ast}\text{\textbar}=k$, such that the empirical risk
tends to the true risk,
$\risk{\text{emp}}{\ermparams{|\window|}{\ast}}\rightarrow\risk{}{\ermparams{|\window|}{\ast}}$. 
To obtain a set of parameters that are a good approximation of the optimal parameters, such that
$\ermparams{|\window|}{\ast}\rightarrow\ermparams{0}{}$, the representational capacity of the model must be increased to
encapsulate more complex underlying
distributions in the data stream~\cite{Cherkassky2007}. In turn, increasing the representational capacity of the model increases the number of
parameters in $\ermparams{|\window|}{\ast}$ which must be learnt. However, by increasing the complexity of the model, more
training samples are required in order to ensure that
the empirical risk, $\risk{\text{emp}}{\ermparams{|\window|}{\ast}}$, is a good approximation of the true risk,
$\risk{}{\ermparams{|\window|}{\ast}}$~\cite{Niyogi1994}.

This means that to ensure a learnt model generalises well and approximates the model with optimal parameters,
$\ermparams{0}{}$, the number of training samples and the model complexity must
grow as a function of one another to guarantee convergence~\cite{Niyogi1994}.
However, in online settings, it may not be feasible to retain a large number of historical instances. Additionally, due
to the dynamic nature of learning in
online environments, even when a large history of instances can be retained, a concept drift may
be encountered prior to observing sufficient instances to estimate the optimal parameters, $\ermparams{0}{}$. 
In the BOTL framework, the meta-learner is trained on a window of instances $\window
=\{\instance{t-|\window|},\dots,\instance{t}\}$, where $|\window|$ is small~\cite{Mckay2019}. As the data stream
progresses, the number of available base models, learnt locally or transferred from other data streams, increases,
which increases the
complexity of the meta-learner, thereby reducing the empirical risk. However, since the window of historical data 
remains small, the likelihood of the meta-learner having poor generalisation
increases as more base models are made available.
Therefore, we consider the conditions under which the probability that the empirical risk,
$\risk{\text{emp}}{\ermparams{|\window|}{\ast}}$, is greater than the true risk, $\risk{}{\ermparams{|\window|}{\ast}}$, plus
some $\epsilon$, is bounded,
\begin{equation}\label{boundrisk}
    \probability{\text{\textbar}\risk{\text{emp}}{\ermparams{|\window|}{\ast}} -
        \risk{}{\ermparams{|\window|}{\ast}}\text{\textbar} > \epsilon}.
\end{equation}

Vapnik~\cite{Vapnik1995} showed that the bound on the empirical risk in Equation~\ref{boundrisk} can be written as
\begin{equation}\label{simpleriskbound}
    \risk{}{\ermparams{|\window|}{\ast}} \leq \risk{\text{emp}}{\ermparams{|\window|}{\ast}} + \varphi(\frac{n}{h_{k}}),
\end{equation}
where $\varphi(\frac{n}{h_{k}})$ is a confidence interval, dependent on the ratio between the number of training samples, $n$,
and the Vapnik-Chervonenkis (VC) dimension, $h_{k}$, which is a measure of the complexity of a model. Therefore, although a complex model may have
a low empirical risk, the
confidence interval may be large when the ratio of training samples to model complexity is large~\cite{Vapnik1995}. In
such cases, a model is said to overfit the training data. In order to reduce the right-hand side of
Equation~\ref{simpleriskbound}, a small confidence interval is required, which would indicate that a model generalises
well on
unseen instances. To minimise the confidence interval, $\varphi(\frac{n}{h_{k}})$, a model with a small VC dimension,
$h_{k}$, must be learnt. However, models
with a small VC dimension have poorer representational capacity, thereby increasing the empirical risk, \risk{\text{emp}}{\ermparams{|\window|}{\ast}}~\cite{Vapnik1995}.
This introduces a trade off between the confidence interval and representational capacity of the model. 
In order to obtain a model that generalises well, we must simultaneously find the VC dimension that minimises the
confidence interval, and the parameter values that minimise the empirical risk~\cite{Cherkassky2007,Niyogi1994,Vapnik1995}. 

To identify the number of base models that should be used, the meta-learner must repeatedly solve this
optimisation problem every time a concept drift is encountered, or a model is received via online TL. For many online
learning applications, solving this optimisation problem is not possible, or practical, due to its computational
complexity. Therefore, alternative approaches to decrease the complexity of the meta-learner must be considered in order
to reduce the confidence interval, such as selecting a subset of the available base models to be used as input
to the meta-learner.


\subsection{Improving Generalisation for Meta-Learners in Online Environments}

Overfitting caused by increasing the representational capacity of a meta-learner is known as the curse of
dimensionality~\cite{Friedman1997}. If future concepts were known, the curse of dimensionality could be avoided by
discarding base models that are known not to be beneficial to the meta-learner for current and future concepts.
However, in online environments future concepts are not known prior to learning. This means that subsets of
base models must be selected by repeatedly evaluating the set of models learnt locally, and transferred when using
online TL, as the data stream progresses.

In online ensembles, one approach to solving
this challenge is to only use the $k$ most recently learnt models as
base models in the ensemble~\cite{Gomes2017,Kolter2005}. However, in many real-world environments that encounter
recurring concepts, recency may not be a good indicator of usefulness~\cite{Gomes2017}.
Instead, other
feature selection and ensemble pruning approaches must be used to determine which models may be most beneficial.
Ensemble pruning is based on the principle that combining the predictions of an appropriate subset of base models will
provide improved
predictive capabilities over combining all base models, as exemplified in bagging and boosting
offline ensembles~\cite{Zhou2012}. Therefore, ensemble pruning techniques can be used to select a subset of base
models for meta-learners in online environments.
Feature selection techniques can be applied by considering the predictions of each base model as input
features to the meta-learner. These meta-features can be evaluated to consider how useful they are for predicting the
current concept~\cite{Kira1992}. 

In an ensemble of regressors there is a bias-variance-covariance trade-off~\cite{Brown2005,Ueda1996}, where the
generalisation error of an ensemble of equally weighted regressors is:
\begin{equation}\label{bvcdecomp}
    \expected{}{(\metamodel{\modelset}-\tLabel{})^2} = \bias^{2} + \frac{1}{M}\var +
\left(1-\frac{1}{|\modelset|}\right)\covar,
\end{equation}
where
\begin{displaymath}
    \begin{split}
    \bias &= \frac{1}{|\modelset|}\sum_{i}\left(\expected{}{\model{}{i}}-\tLabel{}\right)\text{,}\\
    \var &= \frac{1}{|\modelset|}\sum_{i}\expected{}{\model{}{i}-\expected{}{\model{}{i}}}^{2}\text{, and}\\
    \covar &= \frac{1}{|\modelset|(|\modelset|-1)}\sum_{i}\sum_{i\neq
        j}\expected{}{\model{}{i}-\expected{}{\model{}{i}}}\left(\model{}{j}-\expected{}{\model{}{j}}\right).
\end{split}
\end{displaymath}
This bias-variance-covariance decomposition also holds for non-uniformly weighted ensembles~\cite{Brown2005}, and therefore
the bias and variance of base models, and the covariance between them, can impact the generalisation ability of a
meta-learner when base models are not equally weighted. Factors such as these can be considered by evaluating base
models
using metrics to determine which models should be used. Equation~\ref{bvcdecomp} indicates that base models
should be relevant yet diverse in order to prevent the meta-learner overfitting. 

\subsection{Evaluating Base Models}
When undertaking ensemble pruning in offline settings, it is desirable to obtain a relevant yet diverse subset of
base models~\cite{Brown2005}. However, within non-stationary environments base model
diversity is rarely accounted for.
The performance and diversity of model predictions can be used to cull base models~\cite{Mckay2020}. Using the $R^2$
predictive performance of each base model on the current window of data available to the meta-learner allows models
that make poor predictions to be removed. However, using performance alone as a metric to cull base
models may not sufficiently reduce the number of models, allowing the meta-learner to overfit when the
number of potential base models is high~\cite{Mckay2020}. 
The covariance term, $\covar$, in Equation~\ref{bvcdecomp}, accounts for the pairwise difference of base
models~\cite{Ren2016}, which
relates to
their diversity. As the number of base models in the ensemble, $|\modelset|$, increases, the generalisation error
decreases due to the variance term, $\var$. However, increasing the number of base models can cause the covariance
term, $\covar$, to also increase. When simply using the performance of base models as a culling metric, the reduction of
the covariance term in Equation~\ref{bvcdecomp} is not considered. To prevent the covariance term from significantly
increasing, base models must be selected that have small, or negative
covariance~\cite{Ren2016}.

To account for the covariance term in Equation~\ref{bvcdecomp}, the diversity of base models can be used to remove
those that exhibit high covariance. To achieve this, the pairwise MI can be measured between the predictions of each of the base models
on the current window of data available to the meta-learner. When a pair of base models have high MI, the base model with the lower predictive performance
can be culled from the model
set~\cite{Mckay2020}. This reduces the number of redundant models used by the meta-learner, since including models
with similar predictions provides no additional information and increases the covariance term in
Equation~\ref{bvcdecomp}. 

Measuring diversity through
the level of disagreement between base model predictions is a common approach in existing online ensemble pruning
research~\cite{Gomes2017}.
However, this means that the
diversity of base models must be recalculated as new instances are observed in the data stream. 
Although diversity can be estimated using these metrics~\cite{Dutta2009}, it does not
guarantee that the
underlying distributions of data from which each base model was created are diverse~\cite{Dutta2009}. 
Additionally, disagreement in a
regression setting can be highly skewed by a small number of differing predictions made by base models that have
been
learnt from conceptually similar distributions of data. Instead, diversity among the concepts learnt by each base
model can be considered by
estimating conceptual similarity.

Due to concept drift,
base models must be re-evaluated to determine their relevancy to the current concept. Using metrics such as MI to
indicate diversity, which are also dependent on the underlying distribution of observable data, may be undesirable as
repeated pairwise comparisons between base models can become computationally expensive when the number of base
models is large, or
when drifts are encountered frequently. Therefore, the use of a diversity metric, such as conceptual similarity, that
remains static in online environments, even in the presence of concept drift, is beneficial.

\section{Experimental Set-Up}\label{setup}

To evaluate the effectiveness of using conceptual similarity as a metric for base model selection techniques, we use
parameterised thresholding and conceptual clustering to obtain a subset of
base models for the OLS meta-learner in the BOTL framework. RePro~\cite{Yang2005},
ADWIN~\cite{Bifet2007} and AWPro~\cite{Mckay2020} are used as the underlying CDDs to obtain Support Vector Regressors (SVRs) as base models from the
concept drifting data streams. We have chosen SVRs to create base models since they have the representational capacity to
model the underlying concepts encountered in each of the different data stream types used in this paper. However, other
regression models can be used since both the BOTL framework and base model selection techniques are model agnostic. In
order to estimate the conceptual similarity between base models, when a model is selected for transfer in the BOTL
framework, we transfer both the model, $\model{}{i}$, which has been learnt to represent the current concept, and the
reduced PCs, $\pcs{i}$, obtained through Definition~\ref{PAs} and Equation~\ref{selectPCs}. This means that the PCs
associated with each base model are available in each data stream.
We use the datasets provided in~\cite{Mckay2019} and~\cite{Mckay2020}, which include two synthetic data stream
generators, namely the drifting hyperplane data generator~\cite{Mckay2020} and the smart home heating
simulator~\cite{Mckay2019}, and real-world following distance datasets predicting Time To Collision (TTC) from vehicle
telemetry~\cite{Mckay2020}. BOTL is used to
transfer base models between online data streams and CDDs of the same type, such that knowledge is shared between 5
drifting hyperplane, 5 heating simulator, and up to 17 following distance data streams
for each CDD respectively.


\subsection{Baseline Approaches}
To empirically evaluate the effectiveness of estimating conceptual similarity as a diversity metric for selecting a
subset of
base models, we considered using existing ensemble pruning and meta-learner model selection techniques as baseline
approaches. However, most online ensembles that can be used in regression settings, such as AWE~\cite{Wang2003},
OWE~\cite{GomesSoares2015} and AddExp~\cite{Kolter2005}, combine base models using weighted averaging, where weights are
bound between 0 and 1~\cite{Zhou2012}. The use of
bounded weights introduces an assumption that the response variables each base model was learnt to predict have a
consistent range of values. This assumption may not be valid when learning online in real-world environments. For
example, when predicting the desired heating temperature for a smart home heating system, base models learnt over summer
months may have different ranges of response variables in comparison to base models learnt over winter months. Since the
future distribution of response variables is unknown in an online data stream, the response variable cannot be
normalised to ensure all base models are learnt over consistent ranges of response variables. Therefore, bounding base
model weights to be between 0 and 1 may lead to inaccurate predictions. Instead, we consider the underlying techniques
employed by existing online ensembles to prune base models, and use an OLS meta-learner to combine base
models since the weights are not bound between 0 and 1. 
For example AWE~\cite{Wang2003} and AddExp~\cite{Kolter2005} prune base models
using their predictive performance on the current window of observable data, therefore a similar technique, which
evaluates the predictive performance of base models, can be used to
obtain a subset of base models to be input to the OLS meta-learner.

We use the BOTL framework and two existing variants of BOTL that implement base model culling strategies as baseline
approaches~\cite{Mckay2019,Mckay2020}.
We denote frameworks with no base model selection or culling as BOTL,
frameworks with performance thresholding as P-Thresh\footnote{Originally denoted as BOTL-C.I in~\cite{Mckay2019}
    and~\cite{Mckay2020}.}, and frameworks with performance and MI thresholding as MI-Thresh\footnote{Originally denoted
    as BOTL-C.II in~\cite{Mckay2019} and~\cite{Mckay2020}.}.
BOTL uses all transferred models as base models for the OLS meta-learner, whereas P-Thresh and MI-Thresh use na\"{i}ve
culling thresholds that cull base models using similar techniques to pruning strategies implemented by existing online
ensemble approaches. This allows a subset of base models to be selected by removing transferred
models that are expected to be least beneficial.
P-Thresh achieves this by culling transferred
models that obtain $R^{2}$ predictive performances below a threshold, $\thresholdCull{perf}$, on the current window of
observable data, and thus are deemed to be least relevant to the meta-learner. MI-Thresh culls the number of transferred
models more aggressively, by considering the relevancy and diversity of the transferred models. This is achieved through
culling redundant transferred models by considering the pairwise MI between model predictions. For a pair of
models that obtain a MI above a threshold, $\thresholdCull{MI}$, the poorer performing model of the two is culled.
Of the remaining models, those that obtain an $R^{2}$ performance below a performance threshold,
$\thresholdCull{perf}$, are also culled in order to obtain a relevant yet diverse subset of base models.

In addition to the BOTL variants, we also use the underlying CDD alone as a default baseline for what can be achieved
without the use of a meta-learner to combine base model predictions. This allows the benefits and drawbacks of using
meta-learners with differing base model selection techniques to be considered, and highlights the importance of
base model selection strategies when the number of base models becomes large, which can be impacted by the choice of the
underlying CDD.



\subsection{CDDs}
We use RePro~\cite{Yang2005}, ADWIN~\cite{Bifet2007} and AWPro~\cite{Mckay2020} as underlying CDDs to detect concept
drifts and create base models in each data stream. RePro and AWPro
prioritise the reuse of historical models in the presence of recurring concepts~\cite{Mckay2020,Yang2005}. This is
beneficial for online TL since it reduces the communicational overhead of
transferring multiple models that represent the same concept, and also benefits meta-learners by reducing the
number of base models that must be evaluated and compared in order to obtain a relevant yet diverse subset of base models.
Unlike RePro, ADWIN and AWPro estimate the
precise drift point within the window of recent observations so that instances
belonging to the previous concept can be discarded, and do not influence the model learnt to represent a newly
encountered concept~\cite{Bifet2007,Mckay2020}. Discarding instances belonging to the previous concept is
beneficial when evaluating base models since the retention of instances belonging to another concept may influence measures
of diversity.


\subsection{Datasets}\label{datasetsSection}
\setlength{\rotFPtop}{0pt plus 1fil}
\setlength{\rotFPbot}{0pt plus 1fil}
\begin{sidewaystable}
        \setlength{\tabcolsep}{4pt}
    \caption{Dataset characteristics}
    \label{dataSummary}
    \centering
    \begin{tabular}{l|p{4.5cm}ccccc}
        \toprule
        \multirow{2}{*}{Dataset} & \multirow{2}{*}{Characteristic} & \multirow{2}{*}{Dataset Type} & \multicolumn{1}{c}{Drift Type} & \multirow{2}{*}{\#
            Streams} & \multirow{2}{2.2cm}{\centering Avg. \#$\instance{t}$ per \\
            Stream} & \multirow{2}{3cm}{\centering Artificial Drifts\\ per Stream}\\
        & &  & Sudden/Gradual & & & \\
        \midrule
        SuddenA      & Uniform Noise                & Synthetic & \checkmark / --           & 5  & 10,000 & 20 \\ 
        SuddenB      & Sensor Failure               & Synthetic & \checkmark / --          & 5  & 10,000 & 20 \\ 
        SuddenC      & Intermittent Sensor Failure  & Synthetic & \checkmark / --          & 5  & 10,000
        & 20 \\ 
        SuddenD      & Sensor Deterioration         & Synthetic & \checkmark / \checkmark   & 5  & 10,000 & 20 \\ 
        GradualA     & Uniform Noise                & Synthetic &        -- / \checkmark   & 5  &
        11,900 & 20 \\ 
        GradualB     & Sensor Failure               & Synthetic & \checkmark / \checkmark   & 5  &
        11,900 & 20 \\ 
        GradualC     & Intermittent Sensor Failure  & Synthetic & \checkmark / \checkmark   & 5  &
        11,900 & 20 \\ 
        GradualD     & Sensor Deterioration         & Synthetic &        -- / \checkmark   & 5  &
        11,900 & 20 \\ 
        Heating      & Weather Data                 & Hybrid   & \checkmark / \checkmark   & 5  & 17,664 & NA \\ 
        Following    & Vehicular Data               & Real-World   & \checkmark / \checkmark   & 17 & 1909 & NA \\ 
        \bottomrule
    \end{tabular}
\end{sidewaystable}
We evaluate base model selection using three dataset types presented in~\cite{Mckay2019} and~\cite{Mckay2020}, namely
drifting hyperplane, smart home heating
simulator, and following distance datasets. The drifting hyperplane datasets are synthetic datasets that allow sudden,
gradual and recurring drifts to be artificially introduced into the data streams. Common challenges of learning in online
environments have been simulated in the drifting hyperplane datasets, including sensor failure, intermittent sensor failure,
and sensor deterioration. 
The smart home heating simulator uses
real-world weather data and a synthetic heating schedule to create a dataset of desired heating temperatures. Using
real-world weather data to generate this dataset enables base
model selection techniques to be evaluated on data streams containing
drifts that are more representative of those seen in real-world environments. 
Finally, the real-world following distance
dataset predicts Time To Collision (TTC) from vehicle telemetry. An overview of the characteristics of these datasets is
presented in Table~\ref{dataSummary}. 



\subsubsection{Drifting Hyperplane}
\sloppy The drifting hyperplane datasets are modifications of a commonly used benchmark data\-set~\cite{Kolter2005}, adapted for
regression settings~\cite{Mckay2020}.
Each instance at time $t$, $\instance{t}$, is a vector, 
$\instance{t} = \{ \instance{t_{1}},\instance{t_{2}}, \dots,\instance{t_{n}}\}$, containing $n$ randomly generated,
uniformly distributed, variables, $\instance{t_{n}} \in[0,1]$. For each instance, $\instance{t}$, a response variable,
$\tLabel{t}\in[0,1]$, is created using the function 
$\tLabel{t} =
(\instance{t_{p}}+\instance{t_{q}}+\instance{t_{r}})/3$, where $p$, $q$, and $r$
reference three of the $n$ variables of
instance $\instance{t}$. This function represents the underlying concept, $c_{a}$, to be learnt and predicted. Concept
drifts are introduced by modifying which features are used to create $\tLabel{t}$. For example,
an alternative concept, $c_{b}$, may be represented by the function $\tLabel{t} =
(\instance{t_{u}}+\instance{t_{v}}+\instance{t_{w}})/3$, where 
$\{p,q,r\} \neq \{u,v,w\}$ such that $c_{a} \neq c_{b}$. 


A variety of drift types have been synthesised in this generator, including sudden drifts, gradual drifts and recurring
drifts. A sudden
drift from concept $c_{a}$ to concept $c_{b}$ is created between time steps $t$ and $t+1$ by instantaneously
changing the underlying function used to create $\tLabel{t}$ to an alternative function for $\tLabel{t+1}$. A gradual
drift from concept $c_{a}$ to $c_{b}$ occurs between time steps
$t$ and $t+m$, where $m$ instances of data are observed during the drift.
Instances of data created between $t$ and $t+m$ use one of the underlying concept functions, $c_{a}$ or $c_{b}$, to
determine the response variable.
The probability of an instance belonging to concept $c_{a}$ decreases proportionally to the number of instances
seen after time $t$, while the probability of it belonging to $c_{b}$ increases proportionally as we
approach $t+m$. Recurring drifts are created by introducing a concept
$c_{c}$ that reuses the underlying function defined by a previous concept, $c_{a}$, such that we achieve conceptual
equivalence where $c_{c}=c_{a}$.

We create four variations of the drifting hyperplane datasets, introducing concept drifts that represent different
problems that may be encountered when learning in real-world environments. The first variation
simply introduces uniform noise, where $\tLabel{t}\pm0.05$ with probability $0.2$.
Datasets generated in this way are denoted as SuddenA and GradualA for sudden and gradual
drifting data streams respectively.  The second variant simulates sensor failure by setting a feature vector, $i$, to
$0$ at time $t$ for the remainder of the
data stream with probability $0.001$, such that $\instance{t_{i}} = 0$. In the scenario where feature $i$ is used to create
the response variable $\tLabel{}$, we modify two other randomly selected feature vectors, $j$ and $k$, such that $\instance{t_{j}} =
\instance{t_{i}}/4$ and $\instance{t_{k}} = 3\instance{t_{i}}/4$. This ensures that the underlying concept can still be
learnt from the data. We denote datasets generated in this way as SuddenB and GradualB for sudden and gradual
drifting data streams respectively. 

The third variation simulates intermittent sensor failure by selecting a feature vector $i$ to fail at time $t-1$ with
probability $0.001$. Once selected to fail, the feature
value at all subsequent time steps $t$ is set to $0$ such that  $\instance{t_{i}} = 0$
with probability $0.3$. Datasets generated using this scenario are denoted as SuddenC and GradualC.

The final variant emulates the deterioration of a
sensor by including noise depending on the time step $t$, such that $\instance{t_{i}}
= \instance{t_{i}} \pm (0.2(t/|\datastream{}|))$, where $0.2$ is the maximum amount of noise added to 
$\instance{t_{i}}$ and $|\datastream{}|$ is the number of instances in the dataset. This means that as the data stream
progresses, more noise is added to an individual feature, simulating the gradual deterioration in accuracy of a sensor
over time. Additionally, the probability of a sensor deteriorating increases as the data stream progresses, such that the
probability of a feature being selected for deterioration at time $t$ is $0.001(t/|\datastream{}|)$. Datasets generated
in this way are denoted as SuddenD and GradualD for sudden and gradual drifting data streams respectively.

\subsubsection{Heating Simulator}
This dataset is generated from a simulation of a smart home heating system, 
determining the desired room temperature for a user~\cite{Mckay2019}. 
Heating temperatures were derived using weather data collected from a weather station in Birmingham, UK, from
2014 to 2016. This dataset contained rainfall, temperature and sunrise patterns, which were combined with a
schedule, obtained from sampling an individual's pattern of life, to determine when the heating system should
be engaged. The schedule was synthesised to vary the desired temperatures based on time of day, day of week, and
external weather conditions, creating concepts with more complex underlying distributions than those in the drifting
hyperplane datasets. To create
multiple data streams, weather data was sampled from
overlapping time periods and used as input to the synthesised schedule to determine the desired heating temperatures.
Due to the dependencies on weather data, each stream was subject to large amounts
of noise. Concept drifts were introduced manually by changing the schedule, however, drifts
also occurred naturally due to chan\-ging weather conditions. 
By sampling weather data from overlapping time periods, and due to seasonality,
data streams follow similar trends, ensuring the predictive
performance obtained in each data stream can benefit from knowledge transfer. By using concepts that have more complex
underlying distributions, and are dependent on noisy data, the evaluation of BOTL and base model selection techniques on
these data streams is more
indicative of what is achievable when
used in real-world
environments.

\subsubsection{Following Distance} 
Finally, the following distance dataset uses a vehicle's following distance and
speed to calculate Time To Collision (TTC) when following another vehicle~\cite{Mckay2019}. Vehicle
telemetry data such as speed, gear position, brake pressure, throttle position
and indicator status, alongside sensor data that infer external conditions, such as temperature, headlight status,
and windscreen wiper status, were recorded at a sample rate of 1Hz. Additionally, a selection of signals such as vehicle
speed, brake pressure and throttle position were averaged over a window of 5 seconds to capture a recent
history of vehicle state. Vehicle telemetry and environmental data can be used to make predictions that allow 
vehicle functionalities to be personalised and reflect current driving conditions. For example, Adaptive Cruise Control
(ACC) can be personalised by predicting TTC to identify a driver's preferred following distance.
Data was collected from 4 drivers for 17 journeys which varied in duration, collection time and route. Each journey is
considered to be an independent data stream, where 6 data streams were generated by 2 drivers driving 3 pre-defined
routes\footnote{This dataset is available at:
    \url{https://github.com/hmckay/BOTL/blob/master/FollowingDistanceData.zip}}, while the
remaining 11 were generated by 2 drivers
mostly commuting to and from the University of Warwick, UK. The maximum, minimum and average duration
of these journeys were 83 minutes, 15 minutes and 43 minutes respectively. Each data stream is
subject to concept drifts that occur naturally due to changes in the surrounding environment, such as road
types and traffic conditions. BOTL enables knowledge to be learnt and transferred across journeys and between
drivers. 


\section{Experimental Results}\label{results}
\setlength{\tabcolsep}{1.5pt}

To evaluate the use of conceptual similarity to obtain a relevant yet diverse subset of base models, we use our base
model selection techniques for the OLS meta-learner in BOTL and compare these model selection techniques to using the
underlying CDD alone, and to BOTL without base
model selection. We consider using performance thresholding (P-Thresh), MI thresholding (MI-Thresh), conceptual
similarity thresholding (CS-Thresh), and conceptual similarity clustering (CS-Clust) as methods to obtain subsets of
base models. We compare the predictive performances of each of these techniques and also consider the number of
relevancy and diversity metric calculations required to obtain the subsets, which includes the pairwise comparisons
and evaluations of
the predictive performances of base models.

We evaluate these approaches using the datasets originally introduced to evaluate the BOTL
framework~\cite{Mckay2019,Mckay2020}, namely the sudden and gradual drifting hyperplane, smart home heating simulator, and
following distance datasets, as detailed in Section~\ref{datasetsSection}.
Window sizes of 30 instances for the sudden and gradual drifting hyperplane data streams, 480 instances
capturing 10 days of heating and weather observations for smart home heating simulator, and 90 instances
capturing 90 seconds of vehicle telemetry for following distance data streams, are used by the 
underlying CDDs and meta-learner\footnote{For code and reproducibility notes see
    \url{https://github.com/hmckay/BOTLv2}}.
For CS-Thresh and MI-Thresh, we used a
performance threshold, $\thresholdCull{\text{perf}}=0.2$, to ensure that diverse base models are also relevant to the current
concept. This performance threshold value has been chosen for MI-Thresh and CS-Thresh based on the
results obtained by P-Thresh, presented in Figures~\ref{paramsweepsSudden}--\ref{paramsweepsFollow}, and is consistent
with performance thresholds used in~\cite{Mckay2020}.
We also consider using conceptual clustering prior to knowledge transfer to reduce
the number of base models transferred between data streams.


\subsection{Parameterised Culling Thresholds}
\begin{figure}[tbp]
\centering
\centerline{\includegraphics[width=0.95\columnwidth]{{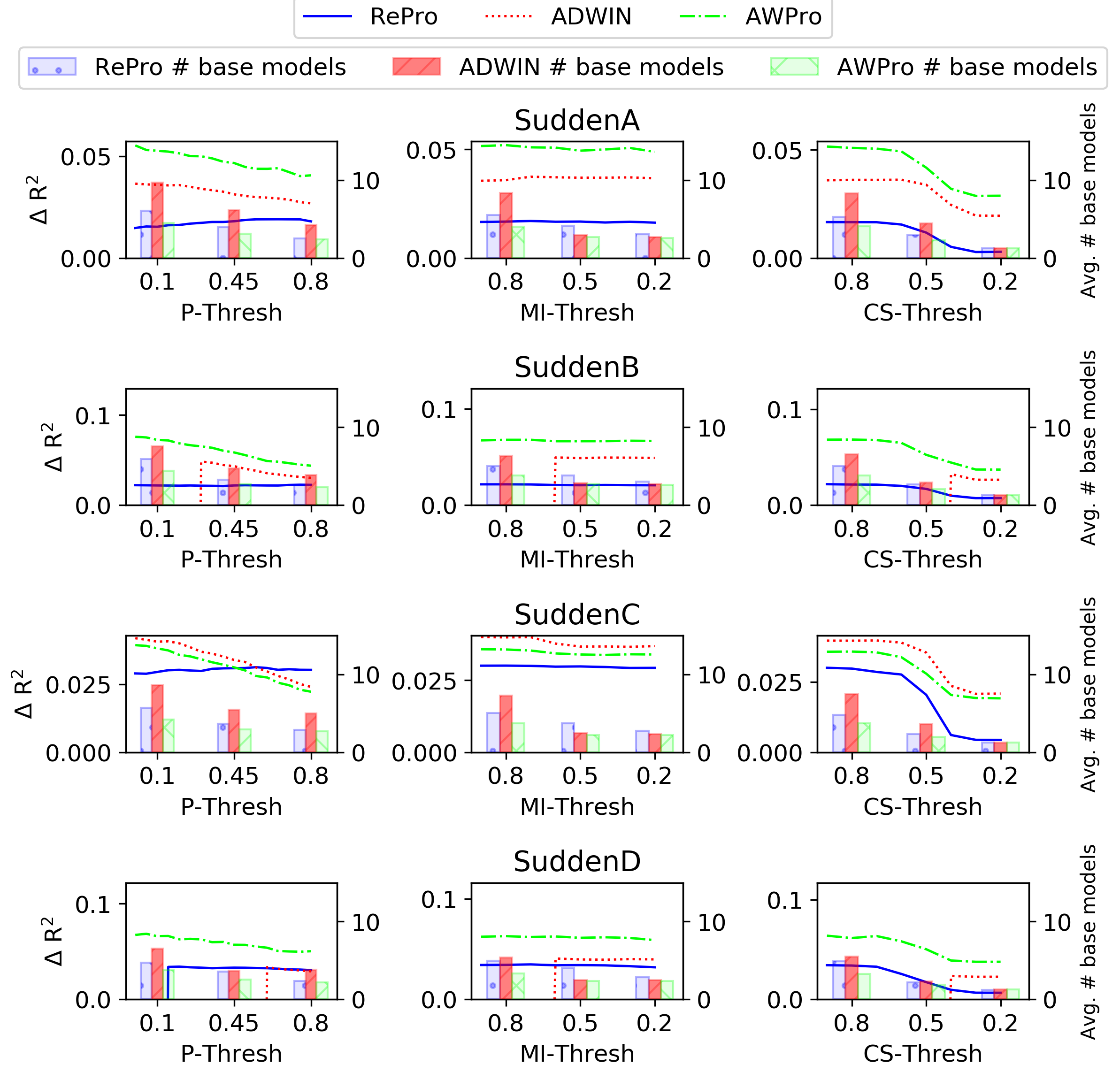}}}
\caption[Caption test]{Sudden Drifting Hyperplane: increase in $R^2$ performance compared to using the underlying CDD
    alone, and number of base models used
    by the BOTL meta-learner using increasingly aggressive culling threshold parameter
    values
    for performance, MI, and conceptual
    similarity thresholding, for variants of the sudden drifting hyperplane datasets when transferring base models between 5 data streams.}
\label{paramsweepsSudden}
\end{figure}
\begin{figure}[tbp]
\centering
\centerline{\includegraphics[width=0.95\columnwidth]{{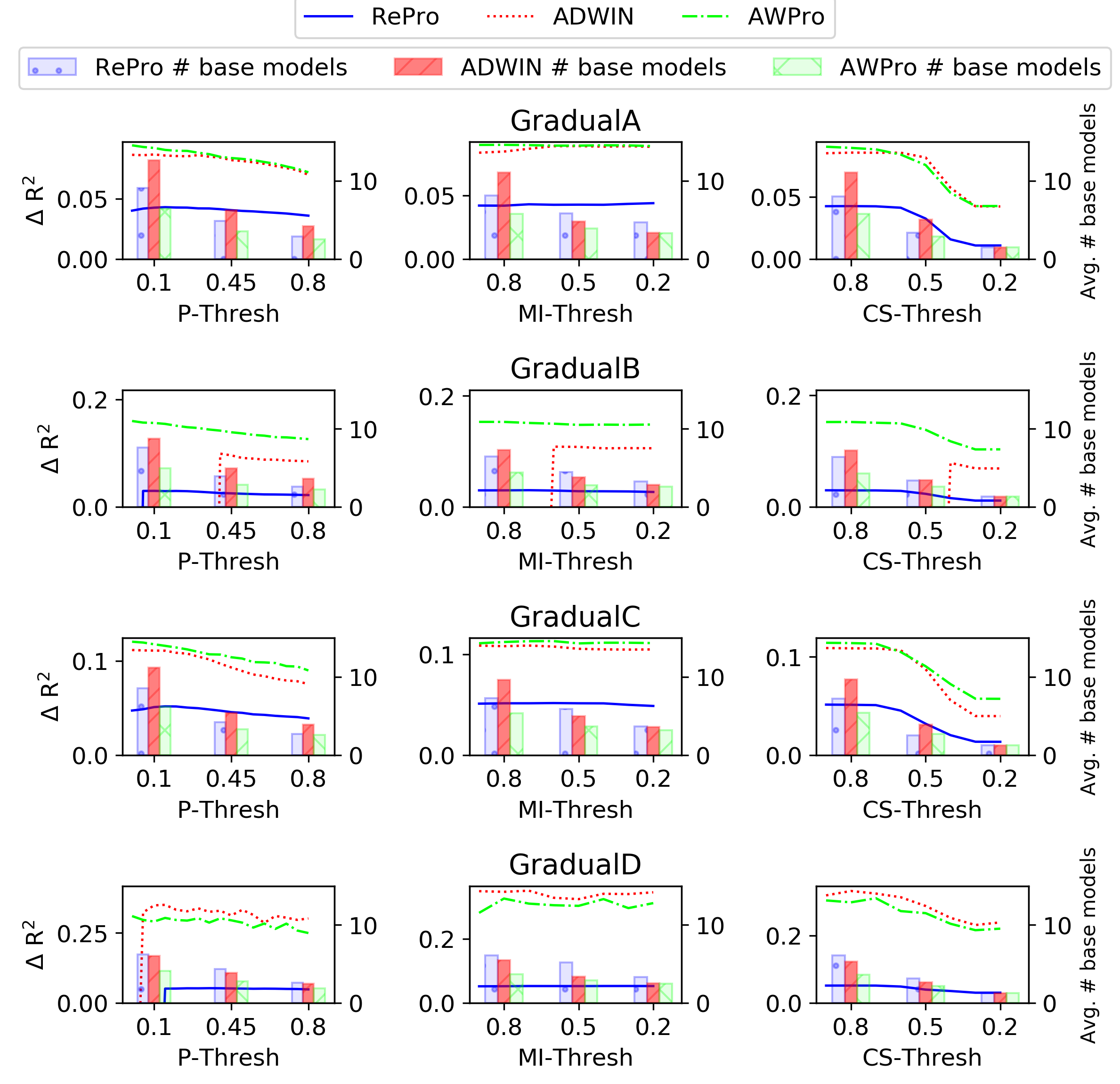}}}
\caption{Gradual Drifting Hyperplane: increase in $R^2$ performance compared to using the underlying CDD
    alone, and number of base models used
    by the BOTL meta-learner using increasingly aggressive culling threshold parameter values for performance, MI, and conceptual
    similarity thresholding, for variants of the gradual drifting hyperplane datasets when transferring base models between 5 data streams.}
\label{paramsweepsGradual}
\end{figure}

\begin{figure}[tbp]
\centering
\centerline{\includegraphics[width=0.95\columnwidth]{{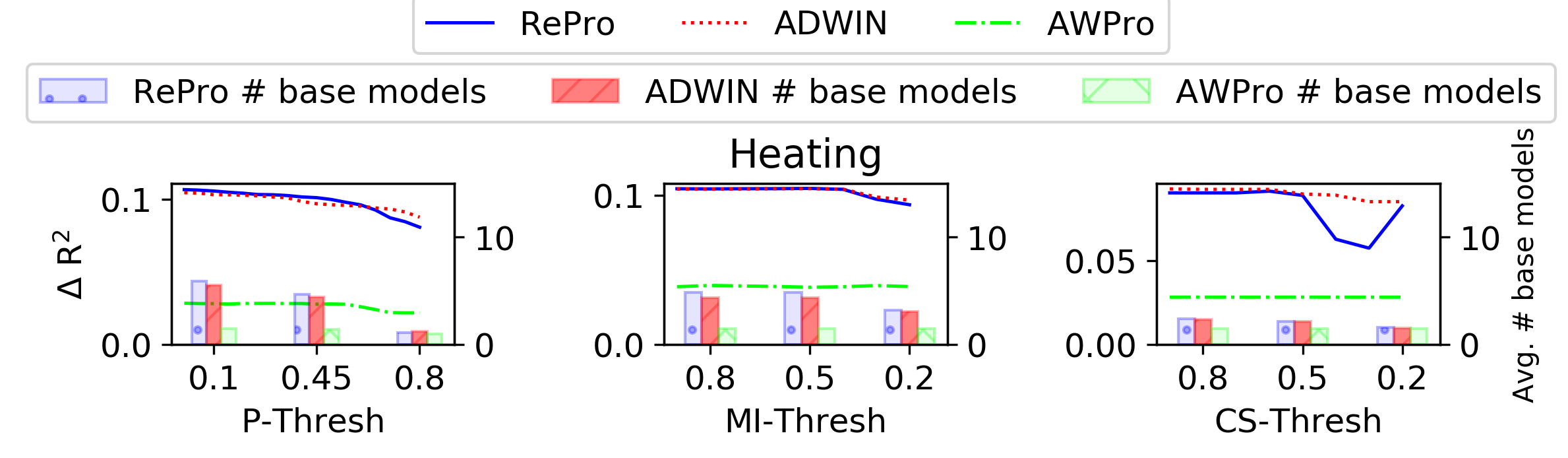}}}
\caption{Heating Simulator: increase in $R^2$ performance compared to using the underlying CDD alone, and
    number of base models used
    by the BOTL meta-learner using increasingly aggressive culling threshold parameter values for performance, MI, and conceptual
    similarity thresholding, for the smart home heating simulator datasets when transferring base models between 5 data streams.}
\label{paramsweepsHeat}
\end{figure}

\begin{figure}[tbp]
\centering
\centerline{\includegraphics[width=0.95\columnwidth]{{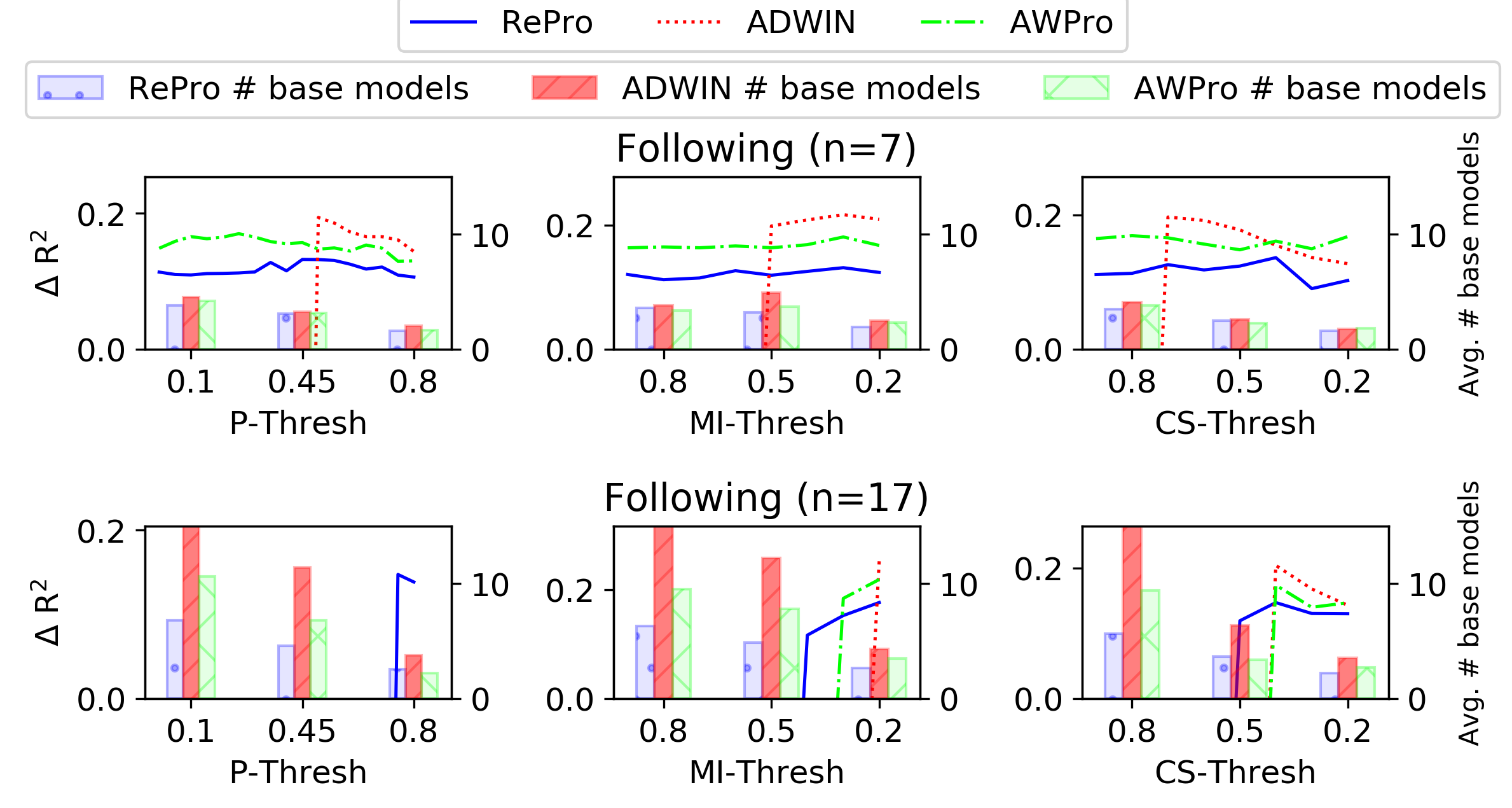}}}
\caption{Following Distance: increase in $R^2$ performance compared to using the underlying CDD alone, and
    number of base models used
    by the BOTL meta-learner using increasingly aggressive culling threshold parameter values for performance, MI, and conceptual
    similarity thresholding, for following distance datasets when transferring base models between 7 and 17 data streams.}
\label{paramsweepsFollow}
\end{figure}

Figures~\ref{paramsweepsSudden}--\ref{paramsweepsFollow} show the
increase in performance of the BOTL meta-learner compared to using the underlying CDD alone, with increasingly aggressive
culling parameters\footnote{For
        Figures~\ref{paramsweepsSudden}--\ref{paramsweepsFollow},
        all plots show increasingly aggressive culling parameter values on the x-axis with the least aggressive
        parameter value on the left of each plot, and most aggressive on the right.} 
for the sudden drifting hyperplane variants, gradual drifting hyperplane variants, smart home heating
simulator and
following data datasets respectively. Analysing the performance of the meta-learner with varying culling
parameter values highlights that the selection of such parameter values is challenging, and can be dependent on many
underlying factors, including the number of base models available, noise in the data stream, separability of base
models for a given diversity metric, and the underlying CDD. 

For sudden and gradual drifting hyperplane data streams with uniform noise (SuddenA, GradualA), in
Figures~\ref{paramsweepsSudden} and~\ref{paramsweepsGradual}, the meta-learner is unlikely to overfit,
regardless of the number of base models, since the concepts to be learnt are simple and there is little noise in these
variants of the synthetic data streams. This means that using aggressive culling parameters reduces the overall
predictive performance of the meta-learner, since the meta-learner benefits from retaining more base models 
without suffering from 
the curse of dimensionality. 
However,
in drifting hyperplane data streams with sensor failure (SuddenB and Gradual B), more aggressive culling parameters are
required to prevent the meta-learner from overfitting when using ADWIN as the underlying CDD. This is necessary since
the increase in noise increases the likelihood of the meta-learner overfitting, and ADWIN does not
make use of historical models in the presence of recurring concepts, leading to large covariances between
base models that have been learnt to represent the same concept. This indicates that aggressive culling
techniques may be required in noisy data streams when using CDDs that do not reuse previously learnt models when concepts
re-occur. Conversely, CDDs that reuse base models, such as RePro and AWPro, benefit from less aggressive culling
parameters in these synthetic drifting hyperplane data streams.


When learning concepts with more complex underlying distributions, the meta-learner is more likely to overfit
when the number of base models
is large in comparison to the size of the window of available data. Data streams created using the smart home heating
simulator are generated using real-world weather data, meaning that the concepts to be learnt are more
complex, and contain more noise, in comparison to the drifting hyperplane data streams. These factors indicate
that the meta-learner is likely to overfit when the number of base models becomes large, therefore requiring aggressive
culling parameters. However, Figure~\ref{paramsweepsHeat} shows that less aggressive culling parameters can be chosen.
This is observed since a large window size has been used to detect concept drifts, create base models, and train the
meta-learner, in these data streams. This highlights the relationship between
the meta-learners generalisation ability, representational capacity, and the amount of available training data.

In addition to this, a wider variety of culling parameters obtain similar predictive performances in the smart home
heating simulator data streams. For example, there is little change in the predictive
performance and number of base models selected for P-Thresh values ranging between 0.1
and 0.5, MI-Thresh values between 0.8 and 0.4, and CS-Thresh values between 0.8 and 0.5. 
A wider range of values can be used as culling thresholds for base model selection in the smart home heating
simulator data streams because the number of base models available to the meta-learner does not change significantly
over these ranges of culling parameter values. This indicates that the meta-learner's sensitivity to culling parameter values
is also dependent on how separable base models are for a given metric. For example, if large numbers of base models
are equally diverse, aggressive culling parameters may be required to sufficiently reduce the number
of base models available to the meta-learner to prevent overfitting.

Figure~\ref{paramsweepsFollow} shows the increase in predictive performance of the meta-learner compared to using the
underlying CDD alone when transferring base models between 7 and 17 following distance data streams. We present the
results for frameworks with differing numbers of data streams to highlight the difficulty in selecting culling parameter
values when the
number of data streams in the framework is large. The 7 data streams were chosen to be representative of the
dataset containing 17 data streams, and contains vehicle telemetry data from one driver following two pre-defined
routes, while the remaining 5 streams were obtained from two drivers commuting.
The meta-learners
that use ADWIN as the underlying CDD are most sensitive to the culling parameter values due to the increased
covariance between base models in the presence of recurring concepts. However, when base models are transferred
between 17 data streams, all meta-learners become sensitive to culling parameter values, regardless of the underlying
CDD used. To prevent overfitting, aggressive culling parameters are required due to two factors. Firstly, using
bi-directional knowledge transfer between 17 data streams increases the number of base models available to the
meta-learner, and therefore aggressive culling parameters are required to sufficiently reduce the number of base models
used as input to the meta-learner for the given window size of available data. 
Secondly, aggressive culling parameters
are required since there may be high levels of covariance between predictions of base models from different data
streams that encounter similar concepts. These challenges are highlighted when using P-Thresh
as a culling mechanism. P-Thresh selects base models using performance alone, and when selecting base models from 17
following
distance data streams the meta-learner overfits, even with aggressive culling parameters. This occurs because, after
culling, the remaining
base models obtain high $R^2$ performances, indicating similar predictions. This introduces high levels of
covariance between inputs to the meta-learner. MI-Thresh and CS-Thresh also suffer from the increased number of base
models and the covariance among the predictions of base models from different data streams. However, a wider
range of culling parameters can be used to obtain a meta-learner with improved predictive performance over the
underlying CDD. This reiterates the importance of using diversity when selecting base models.

The results presented in this section demonstrate that selecting appropriate culling threshold parameter values is
challenging and may require
domain expertise in order to prevent the meta-learner from overfitting. The aggressiveness of the culling threshold
chosen can be dependent on the amount of training data available to the meta-learner, the number of base models
available to select from, the complexity of the underlying distribution of the data stream, and the diversity between
base models. Since each of these factors must be considered, selecting a culling parameter value may be difficult to
determine in advance, and selecting a single threshold parameter for all online environments is not possible. However,
the results presented in Figures~\ref{paramsweepsSudden}--\ref{paramsweepsFollow} show that aggressive culling
parameters are typically more beneficial since less aggressive
culling parameters can be ineffective at reducing the number of base models sufficiently to prevent the meta-learner
overfitting, as can be seen in the following distance datasets. Additionally, only a
small reduction in predictive performance is observed when using an aggressive culling parameter in comparison to less
aggressive culling parameters that are able to prevent the
meta-learner from overfitting, as seen in the drifting hyperplane and heating simulator datasets. 

Since the selection of effective culling threshold parameters is challenging, parameterless base model selection
techniques are required. This can be achieved using conceptual similarity clustering, as introduced in
Section~\ref{paramConceptualClustering}. 
To compare our parameterless clustering approach to parameterised thresholding, we considered the results presented in
Figures~\ref{paramsweepsSudden}--\ref{paramsweepsFollow}, and
selected parameter values that showed an increase in predictive performance in comparison to using the underlying CDD
alone across all dataset types. Therefore, in the remainder of this paper, we use
MI-Thresh with $\thresholdCull{MI}=0.2$ and CS-Thresh with $\thresholdCull{CS}=0.4$ to compare parameterised culling
thresholds with parameterless clustering. P-Thresh is not used as a baseline technique in the
remainder of this paper since it is not able to effectively obtain a subset of base models to prevent overfitting in the
BOTL framework with 17 following distance data streams.

\subsection{Parameterless Clustering}

\begin{table}[htbp]
\centering
    \begin{subtable}{\columnwidth}
    \caption{SuddenA: sudden drifting hyperplanes with uniform noise.}
    \label{suddenRes}
    \vspace{-0.1in}
    \resizebox{\columnwidth}{!}{
    \begin{tabular}{l|rrrrr|rrrrr|rrrrr}
        \toprule
        \multicolumn{1}{c}{}&\multicolumn{5}{c}{RePro}&\multicolumn{5}{c}{ADWIN}&\multicolumn{5}{c}{AWPro}\\
        \multicolumn{1}{c}{}& \multicolumn{1}{c}{$R^2$} & \multicolumn{1}{c}{PMCC$^2$} & \multicolumn{1}{c}{RMSE} &
        \multicolumn{1}{c}{$|\modelsubset|$} &\multicolumn{1}{c}{M.Calcs.} &
        \multicolumn{1}{c}{$R^2$} & \multicolumn{1}{c}{PMCC$^2$} & \multicolumn{1}{c}{RMSE} &
        \multicolumn{1}{c}{$|\modelsubset|$} &\multicolumn{1}{c}{M.Calcs.} &
        \multicolumn{1}{c}{$R^2$} & \multicolumn{1}{c}{PMCC$^2$} & \multicolumn{1}{c}{RMSE} &
        \multicolumn{1}{c}{$|\modelsubset|$} &\multicolumn{1}{c}{M.Calcs.} \\
        \midrule
CDD         & 0.886 & 0.887 & 0.062 & 1  & 0                    & 0.851 & 0.854 & 0.071 & 1  & 0                    & 0.830 & 0.835 & 0.076 & 1 & 0 \\
BOTL        & 0.831 & 0.843 & 0.076 & 23 & 0                    & 0.825 & 0.837 & 0.077 & 40 & 0                    & *\textbf{0.884} & 0.886 & 0.063 & 16 & 0 \\
MI-Thresh   & 0.902 & 0.902 & 0.058 & 3  & 28507                & 0.887 & 0.888 & 0.062 & 2  & 51788                & 0.879 & 0.880 & 0.064 & 2 & 16699 \\
CS-Thresh   & *0.891 & 0.891 & 0.061 & 1  & 2597                & *0.874 & 0.876 & 0.065 & 1  & 2480                & *0.861 & 0.863 & 0.069 & 1 & 1907 \\
CS-Clust    & *\textbf{0.903} & 0.904 & 0.058 & 4  & 5590       & *\textbf{0.884} & 0.885 & 0.063 & 4  & 5174       & *0.876 & 0.878 & 0.065 & 4 & 1188 \\
CS-ClustRed & *0.901 & 0.901 & 0.058 & 3  & 3876                & *\textbf{0.884} & 0.885 & 0.063 & 4  & 2135       & *0.875 & 0.877 & 0.065 & 3 & 809 \\
        \bottomrule
    \end{tabular}}
\end{subtable}

\begin{subtable}{\columnwidth}
    \caption{SuddenB: sudden drifting hyperplanes with single sensor failure.}
    \label{suddenARes}
    \vspace{-0.1in}
    \resizebox{\columnwidth}{!}{
    \begin{tabular}{l|rrrrr|rrrrr|rrrrr}
        \toprule
        \multicolumn{1}{c}{}&\multicolumn{5}{c}{RePro}&\multicolumn{5}{c}{ADWIN}&\multicolumn{5}{c}{AWPro}\\
        \multicolumn{1}{c}{}& \multicolumn{1}{c}{$R^2$} & \multicolumn{1}{c}{PMCC$^2$} & \multicolumn{1}{c}{RMSE} &
        \multicolumn{1}{c}{$|\modelsubset|$} &\multicolumn{1}{c}{M.Calcs.} &
        \multicolumn{1}{c}{$R^2$} & \multicolumn{1}{c}{PMCC$^2$} & \multicolumn{1}{c}{RMSE} &
        \multicolumn{1}{c}{$|\modelsubset|$} &\multicolumn{1}{c}{M.Calcs.} &
        \multicolumn{1}{c}{$R^2$} & \multicolumn{1}{c}{PMCC$^2$} & \multicolumn{1}{c}{RMSE} &
        \multicolumn{1}{c}{$|\modelsubset|$} &\multicolumn{1}{c}{M.Calcs.} \\
        \midrule
CDD         & 0.883  & 0.884 & 0.059 & 1  & 0                   & 0.830  & 0.836 & 0.072  & 1  & 0                  & 0.807  & 0.813 & 0.076  & 1 & 0 \\
BOTL        & -1e+21 & 0.506 & +3e+9 & 25 & 0                   & -5e+22 & 0.499 & +2e+10 & 40 & 0                  & -2e+22 & 0.534 & +1e+10 & 18 & 0 \\
MI-Thresh   & 0.904  & 0.905 & 0.054 & 3  & 27226               & 0.880  & 0.881 & 0.060  & 2  & 46413              & 0.873  & 0.875 & 0.062  & 2 & 16748 \\
CS-Thresh   & *0.892  & 0.893 & 0.057 & 1  & 2403               & *\textbf{0.862}  & 0.864 & 0.065  & 1  & 3038     & *0.851  & 0.852 & 0.067  & 1 & 2398 \\
CS-Clust    & -1e+17 & 0.899 & +5e+6 & 4  & 5134                & -1e+18 & 0.873 & +1e+7  & 4  & 5415               & *\textbf{0.864}  & 0.866 & 0.064  & 4 & 1369 \\
CS-ClustRed & *\textbf{0.904}  & 0.905 & 0.054 & 4  & 3366      & -8e+17 & 0.842 & +2e+7  & 4  & 2393               & *0.862  & 0.863 & 0.065  & 3 & 841 \\
        \bottomrule
    \end{tabular}}
\end{subtable}

\begin{subtable}{\columnwidth}
    \caption{SuddenC: sudden drifting hyperplanes with intermittent single sensor failure.}
    \label{suddenBRes}
    \vspace{-0.1in}
    \resizebox{\columnwidth}{!}{
    \begin{tabular}{l|rrrrr|rrrrr|rrrrr}
        \toprule
        \multicolumn{1}{c}{}&\multicolumn{5}{c}{RePro}&\multicolumn{5}{c}{ADWIN}&\multicolumn{5}{c}{AWPro}\\
        \multicolumn{1}{c}{}& \multicolumn{1}{c}{$R^2$} & \multicolumn{1}{c}{PMCC$^2$} & \multicolumn{1}{c}{RMSE} &
        \multicolumn{1}{c}{$|\modelsubset|$} &\multicolumn{1}{c}{M.Calcs.} &
        \multicolumn{1}{c}{$R^2$} & \multicolumn{1}{c}{PMCC$^2$} & \multicolumn{1}{c}{RMSE} &
        \multicolumn{1}{c}{$|\modelsubset|$} &\multicolumn{1}{c}{M.Calcs.} &
        \multicolumn{1}{c}{$R^2$} & \multicolumn{1}{c}{PMCC$^2$} & \multicolumn{1}{c}{RMSE} &
        \multicolumn{1}{c}{$|\modelsubset|$} &\multicolumn{1}{c}{M.Calcs.} \\
        \midrule
CDD         & 0.878 & 0.879 & 0.064 & 1  & 0                    & 0.845 & 0.848 & 0.072 & 1  & 0                    & 0.847 & 0.850 & 0.072 & 1  & 0 \\
BOTL        & 0.840 & 0.849 & 0.074 & 24 & 0                    & 0.813 & 0.827 & 0.079 & 40 & 0                    & *\textbf{0.887} & 0.889 & 0.062 & 17 & 0 \\
MI-Thresh   & 0.907 & 0.907 & 0.056 & 2  & 26549                & 0.881 & 0.883 & 0.063 & 2  & 45071                & 0.881 & 0.882 & 0.063 & 2  & 16125 \\
CS-Thresh   & *0.884 & 0.884 & 0.063 & 1  & 2115                & *0.868 & 0.870 & 0.067 & 1  & 2411                & *0.868 & 0.869 & 0.067 & 1  & 1810 \\
CS-Clust    & *\textbf{0.908} & 0.909 & 0.056 & 5  & 5085       & *\textbf{0.883} & 0.884 & 0.063 & 4  & 5175       & *0.882 & 0.883 & 0.063 & 4  & 1244 \\
CS-ClustRed & *0.907 & 0.908 & 0.056 & 4  & 3496                & *\textbf{0.883} & 0.884 & 0.063 & 4  & 2034       & *0.882 & 0.883 & 0.063 & 3  & 846 \\
        \bottomrule
    \end{tabular}}
\end{subtable}

\begin{subtable}{\columnwidth}
    \caption{SuddenD: sudden drifting hyperplanes with gradual sensor deterioration.}
    \label{suddenDRes}
    \vspace{-0.1in}
    \resizebox{\columnwidth}{!}{
    \begin{tabular}{l|rrrrr|rrrrr|rrrrr}
        \toprule
        \multicolumn{1}{c}{}&\multicolumn{5}{c}{RePro}&\multicolumn{5}{c}{ADWIN}&\multicolumn{5}{c}{AWPro}\\
        \multicolumn{1}{c}{}& \multicolumn{1}{c}{$R^2$} & \multicolumn{1}{c}{PMCC$^2$} & \multicolumn{1}{c}{RMSE} &
        \multicolumn{1}{c}{$|\modelsubset|$} &\multicolumn{1}{c}{M.Calcs.} &
        \multicolumn{1}{c}{$R^2$} & \multicolumn{1}{c}{PMCC$^2$} & \multicolumn{1}{c}{RMSE} &
        \multicolumn{1}{c}{$|\modelsubset|$} &\multicolumn{1}{c}{M.Calcs.} &
        \multicolumn{1}{c}{$R^2$} & \multicolumn{1}{c}{PMCC$^2$} & \multicolumn{1}{c}{RMSE} &
        \multicolumn{1}{c}{$|\modelsubset|$} &\multicolumn{1}{c}{M.Calcs.} \\
        \midrule
CDD         & 0.868  & 0.869 & 0.065  & 1  & 0                  & 0.845  & 0.849 & 0.070  & 1  & 0                  & 0.810  & 0.816 & 0.077 & 1  & 0 \\
BOTL        & -9e+21 & 0.340 & +1e+10 & 27 & 0                  & -1e+22 & 0.334 & +1e+10 & 40 & 0                  & -2e+21 & 0.357 & +5e+9 & 16 & 0 \\
MI-Thresh   & 0.900  & 0.900 & 0.056  & 2  & 28310              & 0.885  & 0.886 & 0.061  & 2  & 42263              & 0.873  & 0.874 & 0.063 & 2  & 14352 \\
CS-Thresh   & *0.878  & 0.878 & 0.062  & 1  & 2547              & *0.869  & 0.870 & 0.065  & 1  & 2261              & *0.849  & 0.851 & 0.069 & 1  & 2010 \\
CS-Clust    & *\textbf{0.898}  & 0.898 & 0.057  & 4  & 5647     & *0.882  & 0.883 & 0.061  & 4  & 5144              & *\textbf{0.865}  & 0.866 & 0.065 & 4  & 1194 \\
CS-ClustRed & *0.896  & 0.897 & 0.057  & 4  & 3534              & *\textbf{0.883}  & 0.884 & 0.061  & 4  & 2099     & *0.861  & 0.862 & 0.066 & 3  & 917 \\
        \bottomrule
    \end{tabular}}
\end{subtable}

\caption{Sudden Drifting Hyperplane: $R^2$ , $\text{PMCC}^2$ and RMSE predictive performance, the average number of base
    models used by the meta-learner ($|\modelsubset|$), and the average number of relevancy and diversity metric
    calculations to compare and evaluate base models (M.Calcs.) 
    for variants of the sudden drifting hyperplane datasets when transferring models between 5 data streams in BOTL.
    Improved predictive performances with statistical t-test values $p<0.01$ compared to the underlying CDD, while
    requiring fewer relevancy and diversity metric calculations than MI-Thresh are indicated with $^{\ast}$. Of these,
    bold type indicates the approach with highest performance.}
\label{allSuddenperfs}
\end{table}
\begin{table}[htbp]
\centering
    \begin{subtable}{\columnwidth}
    \caption{GradualA: gradual drifting hyperplanes with uniform noise.}
    \label{gradualRes}
    \vspace{-0.1in}
    \resizebox{\columnwidth}{!}{
    \begin{tabular}{l|rrrrr|rrrrr|rrrrr}
        \toprule
        \multicolumn{1}{c}{}&\multicolumn{5}{c}{RePro}&\multicolumn{5}{c}{ADWIN}&\multicolumn{5}{c}{AWPro}\\
        \multicolumn{1}{c}{}& \multicolumn{1}{c}{$R^2$} & \multicolumn{1}{c}{PMCC$^2$} & \multicolumn{1}{c}{RMSE} &
        \multicolumn{1}{c}{$|\modelsubset|$} &\multicolumn{1}{c}{M.Calcs.} &
        \multicolumn{1}{c}{$R^2$} & \multicolumn{1}{c}{PMCC$^2$} & \multicolumn{1}{c}{RMSE} &
        \multicolumn{1}{c}{$|\modelsubset|$} &\multicolumn{1}{c}{M.Calcs.} &
        \multicolumn{1}{c}{$R^2$} & \multicolumn{1}{c}{PMCC$^2$} & \multicolumn{1}{c}{RMSE} &
        \multicolumn{1}{c}{$|\modelsubset|$} &\multicolumn{1}{c}{M.Calcs.} \\
        \midrule
CDD         & 0.849 & 0.849 & 0.068 & 1  & 0                    & 0.796 & 0.803 & 0.079 & 1  & 0                    & 0.798 & 0.804 & 0.078 & 1 & 0 \\
BOTL        & 0.771 & 0.797 & 0.084 & 30 & 0                    & 0.795 & 0.813 & 0.079 & 38 & 0                    & *\textbf{0.882} & 0.885 & 0.060 & 19 & 0 \\
MI-Thresh   & 0.892 & 0.892 & 0.058 & 3  & 56526                & 0.886 & 0.886 & 0.059 & 3  & 73607                & 0.887 & 0.888 & 0.059 & 3 & 31410 \\
CS-Thresh   & *0.866 & 0.866 & 0.064 & 2  & 3246                & *0.854 & 0.854 & 0.067 & 1  & 2946                & *0.854 & 0.854 & 0.067 & 1 & 2404 \\
CS-Clust    & *\textbf{0.897} & 0.898 & 0.056 & 4  & 11979      & *\textbf{0.877} & 0.878 & 0.061 & 4  & 4985       & *0.876 & 0.877 & 0.062 & 4 & 1448 \\
CS-ClustRed & *0.894 & 0.894 & 0.057 & 4  & 6912                & *0.872 & 0.873 & 0.063 & 3  & 1732                & *0.871 & 0.872 & 0.063 & 3 & 858 \\
        \bottomrule
    \end{tabular}}
\end{subtable}

\begin{subtable}{\columnwidth}
    \caption{GradualB: gradual drifting hyperplanes with single sensor failure.}
    \label{gradualARes}
    \vspace{-0.1in}
    \resizebox{\columnwidth}{!}{
    \begin{tabular}{l|rrrrr|rrrrr|rrrrr}
        \toprule
        \multicolumn{1}{c}{}&\multicolumn{5}{c}{RePro}&\multicolumn{5}{c}{ADWIN}&\multicolumn{5}{c}{AWPro}\\
        \multicolumn{1}{c}{}& \multicolumn{1}{c}{$R^2$} & \multicolumn{1}{c}{PMCC$^2$} & \multicolumn{1}{c}{RMSE} &
        \multicolumn{1}{c}{$|\modelsubset|$} &\multicolumn{1}{c}{M.Calcs.} &
        \multicolumn{1}{c}{$R^2$} & \multicolumn{1}{c}{PMCC$^2$} & \multicolumn{1}{c}{RMSE} &
        \multicolumn{1}{c}{$|\modelsubset|$} &\multicolumn{1}{c}{M.Calcs.} &
        \multicolumn{1}{c}{$R^2$} & \multicolumn{1}{c}{PMCC$^2$} & \multicolumn{1}{c}{RMSE} &
        \multicolumn{1}{c}{$|\modelsubset|$} &\multicolumn{1}{c}{M.Calcs.} \\
        \midrule
CDD         & 0.861  & 0.862 & 0.067 & 1  & 0                   & 0.757  & 0.771 & 0.088 & 1  & 0                   & 0.697  & 0.719 & 0.099 & 1 & 0 \\
BOTL        & -7e+20 & 0.322 & +3e+9 & 29 & 0                   & -1e+21 & 0.322 & +3e+9 & 39 & 0                   & -1e+21 & 0.348 & +3e+9 & 19 & 0 \\
MI-Thresh   & 0.888  & 0.888 & 0.060 & 3  & 45556               & 0.863  & 0.863 & 0.067 & 2  & 58346               & 0.845  & 0.845 & 0.071 & 2 & 23961 \\
CS-Thresh   & *0.876  & 0.877 & 0.063 & 1  & 3497               & *0.834  & 0.835 & 0.073 & 1  & 3186               & *0.814  & 0.815 & 0.078 & 1 & 2554 \\
CS-Clust    & -8e+16 & 0.868 & +8e+6 & 4  & 12941               & *\textbf{0.857}  & 0.858 & 0.068 & 4  & 5189      & *\textbf{0.834}  & 0.835 & 0.073 & 3 & 1481 \\
CS-ClustRed & *\textbf{0.890}  & 0.891 & 0.060 & 3  & 7501      & -4e+15 & 0.855 & +1e+6 & 4  & 2031                & *\textbf{0.834}  & 0.835 & 0.073 & 3 & 925 \\
        \bottomrule
    \end{tabular}}
\end{subtable}

\begin{subtable}{\columnwidth}
    \caption{GradualC: gradual drifting hyperplanes with intermittent single sensor failure.}
    \label{gradualBRes}
    \vspace{-0.1in}
    \resizebox{\columnwidth}{!}{
    \begin{tabular}{l|rrrrr|rrrrr|rrrrr}
        \toprule
        \multicolumn{1}{c}{}&\multicolumn{5}{c}{RePro}&\multicolumn{5}{c}{ADWIN}&\multicolumn{5}{c}{AWPro}\\
        \multicolumn{1}{c}{}& \multicolumn{1}{c}{$R^2$} & \multicolumn{1}{c}{PMCC$^2$} & \multicolumn{1}{c}{RMSE} &
        \multicolumn{1}{c}{$|\modelsubset|$} &\multicolumn{1}{c}{M.Calcs.} &
        \multicolumn{1}{c}{$R^2$} & \multicolumn{1}{c}{PMCC$^2$} & \multicolumn{1}{c}{RMSE} &
        \multicolumn{1}{c}{$|\modelsubset|$} &\multicolumn{1}{c}{M.Calcs.} &
        \multicolumn{1}{c}{$R^2$} & \multicolumn{1}{c}{PMCC$^2$} & \multicolumn{1}{c}{RMSE} &
        \multicolumn{1}{c}{$|\modelsubset|$} &\multicolumn{1}{c}{M.Calcs.} \\
        \midrule
CDD         & 0.841 & 0.841 & 0.068 & 1  & 0                    & 0.773 & 0.779 & 0.081 & 1  & 0                    & 0.759 & 0.773 & 0.083 & 1 & 0 \\ 
BOTL        & 0.784 & 0.806 & 0.080 & 28 & 0                    & 0.778 & 0.800 & 0.081 & 38 & 0                    & *\textbf{0.876} & 0.879 & 0.060 & 19 & 0 \\ 
MI-Thresh   & 0.890 & 0.890 & 0.057 & 3  & 51948                & 0.878 & 0.878 & 0.060 & 3  & 69534                & 0.872 & 0.873 & 0.061 & 3 & 29224 \\ 
CS-Thresh   & *0.861 & 0.861 & 0.064 & 1  & 3132                & *0.829 & 0.830 & 0.071 & 1  & 2882                & *0.826 & 0.827 & 0.071 & 1 & 2482 \\ 
CS-Clust    & *\textbf{0.891} & 0.892 & 0.057 & 4  & 12289      & *\textbf{0.856} & 0.856 & 0.065 & 4  & 4867       & *0.866 & 0.866 & 0.063 & 4 & 1427 \\ 
CS-ClustRed & *0.888 & 0.889 & 0.057 & 4  & 7151                & *0.854 & 0.855 & 0.065 & 3  & 1943                & *0.855 & 0.856 & 0.065 & 3 & 816 \\ 
        \bottomrule
    \end{tabular}}
\end{subtable}

\begin{subtable}{\columnwidth}
    \caption{GradualD: gradual drifting hyperplanes with gradual sensor deterioration.}
    \label{gradualDRes}
    \vspace{-0.1in}
    \resizebox{\columnwidth}{!}{
    \begin{tabular}{l|rrrrr|rrrrr|rrrrr}
        \toprule
        \multicolumn{1}{c}{}&\multicolumn{5}{c}{RePro}&\multicolumn{5}{c}{ADWIN}&\multicolumn{5}{c}{AWPro}\\
        \multicolumn{1}{c}{}& \multicolumn{1}{c}{$R^2$} & \multicolumn{1}{c}{PMCC$^2$} & \multicolumn{1}{c}{RMSE} &
        \multicolumn{1}{c}{$|\modelsubset|$} &\multicolumn{1}{c}{M.Calcs.} &
        \multicolumn{1}{c}{$R^2$} & \multicolumn{1}{c}{PMCC$^2$} & \multicolumn{1}{c}{RMSE} &
        \multicolumn{1}{c}{$|\modelsubset|$} &\multicolumn{1}{c}{M.Calcs.} &
        \multicolumn{1}{c}{$R^2$} & \multicolumn{1}{c}{PMCC$^2$} & \multicolumn{1}{c}{RMSE} &
        \multicolumn{1}{c}{$|\modelsubset|$} &\multicolumn{1}{c}{M.Calcs.} \\
        \midrule
CDD         & 0.845  & 0.846 & 0.080  & 1  & 0                  & 0.484  & 0.593 & 0.136  & 1  & 0                  & 0.513  & 0.623 & 0.130  & 1  & 0 \\
BOTL        & -2e+22 & 0.147 & +2e+10 & 31 & 0                  & -1e+22 & 0.130 & +1e+10 & 25 & 0                  & -2e+22 & 0.111 & +2e+10 & 16 & 0 \\
MI-Thresh   & 0.898  & 0.898 & 0.065  & 3  & 47278              & 0.827  & 0.828 & 0.079  & 2  & 33773              & 0.816  & 0.817 & 0.082  & 2  & 19221 \\
CS-Thresh   & *0.881  & 0.881 & 0.070  & 1  & 4119              & *0.745  & 0.746 & 0.097  & 1  & 2667              & *0.758  & 0.759 & 0.094  & 1  & 2345 \\
CS-Clust    & *\textbf{0.898}  & 0.899 & 0.065  & 5  & 15555    & *0.761  & 0.762 & 0.091  & 3  & 2452              & *\textbf{0.766}  & 0.767 & 0.089  & 3  & 1089 \\
CS-ClustRed & *0.894  & 0.895 & 0.066  & 4  & 9224              & *\textbf{0.766}  & 0.768 & 0.090  & 3  & 1212     & *0.764  & 0.765 & 0.090  & 3  & 704 \\
        \bottomrule
    \end{tabular}}
\end{subtable}
\caption{Gradual Drifting Hyperplane: $R^2$ , $\text{PMCC}^2$ and RMSE predictive performance, the average number of base
    models used by the meta-learner ($|\modelsubset|$), and the average number of relevancy and diversity metric
    calculations to compare and evaluate base models (M.Calcs.) 
    for variants of the gradual drifting hyperplane datasets when transferring models between 5 data streams in BOTL.
    Improved predictive performances with statistical t-test values $p<0.01$ compared to the underlying CDD, while
    requiring fewer relevancy and diversity metric calculations than MI-Thresh are indicated with $^{\ast}$. Of these,
    bold type indicates the approach with highest performance.}
\label{allGradualperfs}
\end{table}

Tables~\ref{allSuddenperfs}--\ref{allFollowingperfs} show the $R^2$,
$\text{PMCC}^2$ and RMSE predictive performances, the average number of base models used by the meta-learner, and the
average number of relevancy and diversity metric calculations required to compare and evaluate base models
for the various datasets and CDDs considered. These results show that, without base model
selection, BOTL is more likely
to overfit in noisy data streams, as seen in Tables~\ref{suddenARes},~\ref{suddenDRes},~\ref{gradualARes},~\ref{gradualDRes},
and~\ref{allFollowingperfs}. However, using base model selection techniques that obtain a relevant yet diverse subset
of base models reduces the likelihood of overfitting.


\begin{table}[tbp]
\centering
    \resizebox{\textwidth}{!}{
    \begin{tabular}{l|rrrrr|rrrrr|rrrrr}
        \toprule
        \multicolumn{1}{c}{}&\multicolumn{5}{c}{RePro}&\multicolumn{5}{c}{ADWIN}&\multicolumn{5}{c}{AWPro}\\
        \multicolumn{1}{c}{}& \multicolumn{1}{c}{$R^2$} & \multicolumn{1}{c}{PMCC$^2$} & \multicolumn{1}{c}{RMSE} &
        \multicolumn{1}{c}{$|\modelsubset|$} &\multicolumn{1}{c}{M.Calcs.} &
        \multicolumn{1}{c}{$R^2$} & \multicolumn{1}{c}{PMCC$^2$} & \multicolumn{1}{c}{RMSE} &
        \multicolumn{1}{c}{$|\modelsubset|$} &\multicolumn{1}{c}{M.Calcs.} &
        \multicolumn{1}{c}{$R^2$} & \multicolumn{1}{c}{PMCC$^2$} & \multicolumn{1}{c}{RMSE} &
        \multicolumn{1}{c}{$|\modelsubset|$} &\multicolumn{1}{c}{M.Calcs.} \\
        \midrule
CDD         & 0.636 & 0.652 & 2.506 & 1   & 0     & 0.635 & 0.655 & 2.509 & 1 & 0     & 0.597 & 0.621 & 2.636 & 1   & 0 \\
BOTL        & *\textbf{0.730} & 0.737 & 2.154 & 10  & 0     & *\textbf{0.720} & 0.728 & 2.190 & 9 & 0     & *\textbf{0.716} & 0.725 & 2.212 & 10  & 0 \\
MI-Thresh   & 0.709 & 0.716 & 2.238 & 3   & 3190  & 0.707 & 0.715 & 2.244 & 3 & 3790  & 0.690 & 0.701 & 2.307 & 3   & 3792 \\
CS-Thresh   & *0.705 & 0.712 & 2.254 & 3   & 496   & *0.701 & 0.709 & 2.270 & 3 & 594   & *0.686 & 0.694 & 2.326 & 3   & 512 \\
CS-Clust    & *0.709 & 0.715 & 2.241 & 3   & 794   & *0.709 & 0.717 & 2.240 & 3 & 877   & *0.689 & 0.699 & 2.313 & 3   & 1178 \\
CS-ClustRed & *0.707 & 0.714 & 2.248 & 3   & 655   & *0.706 & 0.714 & 2.248 & 3 & 794   & *0.693 & 0.702 & 2.296 & 3   & 1015 \\
        \bottomrule
    \end{tabular}}

\caption{Heating Simulator: $R^2$ , $\text{PMCC}^2$ and RMSE predictive performance, the average number of base
    models used by the meta-learner ($|\modelsubset|$), and the average number of relevancy and diversity metric
    calculations to compare and evaluate base models (M.Calcs.) 
    for smart home hearing simulator datasets when transferring models between 5 data streams in BOTL.
    Improved predictive performances with statistical t-test values $p<0.01$ compared to the underlying CDD, while
    requiring fewer relevancy and diversity metric calculations than MI-Thresh are indicated with $^{\ast}$. Of these,
    bold type indicates the approach with highest performance.}
\label{allHeatingperfs}
\vskip -0.1in
\end{table}

\begin{table}[tbp]
\centering
    \resizebox{\textwidth}{!}{
    \begin{tabular}{l|rrrrr|rrrrr|rrrrr}
        \toprule
        \multicolumn{1}{c}{}&\multicolumn{5}{c}{RePro}&\multicolumn{5}{c}{ADWIN}&\multicolumn{5}{c}{AWPro}\\
        \multicolumn{1}{c}{}& \multicolumn{1}{c}{$R^2$} & \multicolumn{1}{c}{PMCC$^2$} & \multicolumn{1}{c}{RMSE} &
        \multicolumn{1}{c}{$|\modelsubset|$} &\multicolumn{1}{c}{M.Calcs.} &
        \multicolumn{1}{c}{$R^2$} & \multicolumn{1}{c}{PMCC$^2$} & \multicolumn{1}{c}{RMSE} &
        \multicolumn{1}{c}{$|\modelsubset|$} &\multicolumn{1}{c}{M.Calcs.} &
        \multicolumn{1}{c}{$R^2$} & \multicolumn{1}{c}{PMCC$^2$} & \multicolumn{1}{c}{RMSE} &
        \multicolumn{1}{c}{$|\modelsubset|$} &\multicolumn{1}{c}{M.Calcs.} \\
        \midrule
CDD         & 0.545 & 0.564 & 0.603 & 1 & 0               & 0.441  & 0.502 & 0.676 & 1  & 0                 & 0.516 & 0.554 & 0.628 & 1 & 0 \\
BOTL        & *0.653 & 0.685 & 0.533 & 5 & 0              & -2e+15 & 0.400 & +2e+7 & 10 & 0                 & 0.456 & 0.671 & 0.586 & 7 & 0 \\
MI-Thresh   & 0.675 & 0.696 & 0.516 & 2 & 869             & 0.659  & 0.678 & 0.530 & 2  & 1302              & 0.684 & 0.695 & 0.512 & 2 & 768 \\
CS-Thresh   & *0.664 & 0.688 & 0.524 & 2 & 147            & *0.613  & 0.640 & 0.567 & 2  & 144              & *0.665 & 0.677 & 0.528 & 2 & 121 \\
CS-Clust    & *\textbf{0.655} & 0.681 & 0.532 & 2 & 587   & *0.634  & 0.661 & 0.543 & 3  & 481              & *0.670 & 0.682 & 0.526 & 2 & 298 \\
CS-ClustRed & *0.650 & 0.678 & 0.535 & 2 & 577            & *\textbf{0.637}  & 0.662 & 0.543 & 2  & 328     & *\textbf{0.678} & 0.689 & 0.521 & 2 & 264 \\
        \bottomrule
    \end{tabular}}

\caption{Following Distance: $R^2$ , $\text{PMCC}^2$ and RMSE predictive performance, the average number of base
    models used by the meta-learner ($|\modelsubset|$), and the average number of relevancy and diversity metric
    calculations to compare and evaluate base models (M.Calcs.) 
    for following distance datasets when transferring models between 7 data streams in BOTL.
    The $R^2$ and
    $\text{PMCC}^2$ predictive performances and number of base model for other numbers of data streams is shown in
    Figure~\ref{allplots}. 
    Improved predictive performances with statistical t-test values $p<0.01$ compared to the underlying CDD, while
    requiring fewer relevancy and diversity metric calculations than MI-Thresh are indicated with $^{\ast}$. Of these,
    bold type indicates the approach with highest performance.}
\label{allFollowingperfs}
\end{table}

CS-Thresh obtains improved predictive performances in comparison to using the
underlying CDD alone (with statistical significance $p<0.01$). CS-Thresh also frequently outperforms BOTL without base
model selection, and obtains comparable predictive
performances to using MI-Thresh. For most datasets, CS-Clust also achieves this. However, CS-Clust obtains a poor $R^2$
predictive performance on SuddenB data streams when using RePro and ADWIN as underlying CDDs, as shown in
Table~\ref{suddenARes}, and when using RePro for GradualB data streams, as shown in Table~\ref{gradualARes}. This
indicates that CS-Clust is still susceptible to overfitting in data streams with large amounts of noise. Noise in these
data streams prevents the clustering approach from effectively distinguishing between models learnt to represent
different concepts. Conceptual clustering is more challenging in noisy environments because there is more variance in the
underlying distribution of data used to create base models. This can affect the PCs created from the underlying
distribution of data belonging to each concept, and can also increase the number of
PCs required to capture 99.9\% of the variance of the window of data. Capturing the increased noise within the PCs can
reduce the PAs between these windows of synthetic data, making conceptual clustering more challenging due to a reduction
in the separability of base models. To overcome this, fewer PCs could be used to calculate the PAs between the
subspaces, thereby capturing less variance in the window of data used to create base
models. Alternatively, other methods of obtaining orthonormal representations of data that are
more robust to noise, such as Laplacian PCA~\cite{Zhao2007} or Robust PCA~\cite{Xu2010}, can be used~\cite{John2019}.

Although CS-Clust overfits in these synthetic data streams, the $\text{PMCC}^{2}$ performances 
are statistically significantly ($p<0.01$) greater than BOTL with no base model selection, and outperform the
$\text{PMCC}^2$ performance obtained when using the underlying CDDs alone. This is observed since $\text{PMCC}^2$ is
bound between 0 and 1, whereas $R^2$ is unbounded, $(-\infty,1]$. This means that the $R^2$ predictive performance can be highly skewed if
the meta-learner overfits for a small number of instances in the data stream. The observed improvement in $\text{PMCC}^2$
performance indicates that although CS-Clust can be susceptible to overfitting in
these environments, the meta-learner overfits less frequently in comparison to the BOTL meta-learner that uses all base
models. With the exception of these noisy synthetic data streams, CS-Clust obtains $R^2$ performances statistically
significantly ($p<0.01$) greater than the underlying CDDs, and outperforms the BOTL meta-learner with no base model
selection techniques for the majority of CDDs, across all datasets except the heating simulator data streams, as shown in
Table~\ref{allHeatingperfs}. 

\begin{figure*}[tb]
\centering
\centerline{\includegraphics[width=\textwidth]{{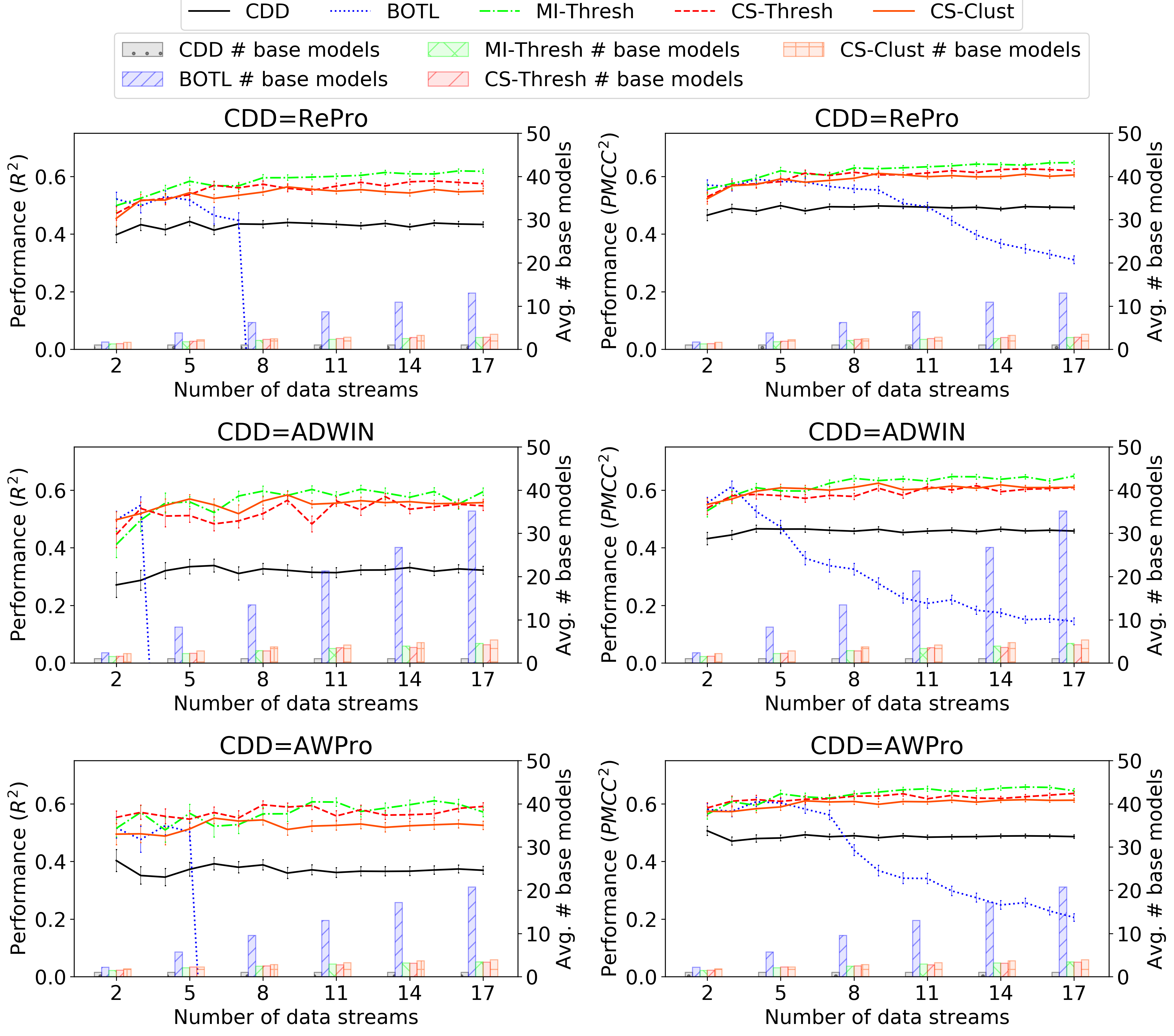}}}
\caption{$R^{2}$ and $\text{PMCC}^2$ predictive performance, and number of base models used by BOTL meta-learners for increasing numbers
of following distance data streams.}
\label{allplots}
\end{figure*}

Table~\ref{allHeatingperfs} shows that the BOTL framework
with no base model selection strategy achieves the highest predictive performance for the smart home heating simulator
data streams. Although the number of base models
used by the meta-learner, $|\modelsubset|$, is greater than the number of base models used by meta-learners in
following distance data streams, shown in Table~\ref{allFollowingperfs}, the BOTL meta-learner does not overfit in the
heating simulator data streams. This is because the window of available data used to learn the meta-learner weights,
$\window$, is larger for the heating simulator data streams (480 instances) in comparison to following distance data streams
(90 instances). This means that the meta-learner is able to use more base models before suffering from the curse of
dimensionality~\cite{Friedman1997}. This shows that base model selection techniques are not always necessary, however,
the number of available base models cannot be known prior to learning. Although BOTL achieves the highest predictive
performance, all base model selection techniques obtain
$R^2$ performances that are statistically significantly ($p<0.01$) greater than the underlying CDD, and obtain similar
predictive performances to BOTL with no base model selection.

Figure~\ref{allplots} shows the predictive performance and number of base models used by the meta-learner with
increasing numbers of following distance data streams. Without using base model selection techniques, BOTL can
overfit, even when the number of data streams is small. However, CS-Thresh prevents the meta-learner overfitting in
these real-world data streams, obtaining similar predictive performances to MI-Thresh, while CS-Clust achieves this
without requiring a user defined threshold parameter.


\begin{figure}[tb]
\centering
\centerline{\includegraphics[width=0.7\columnwidth]{{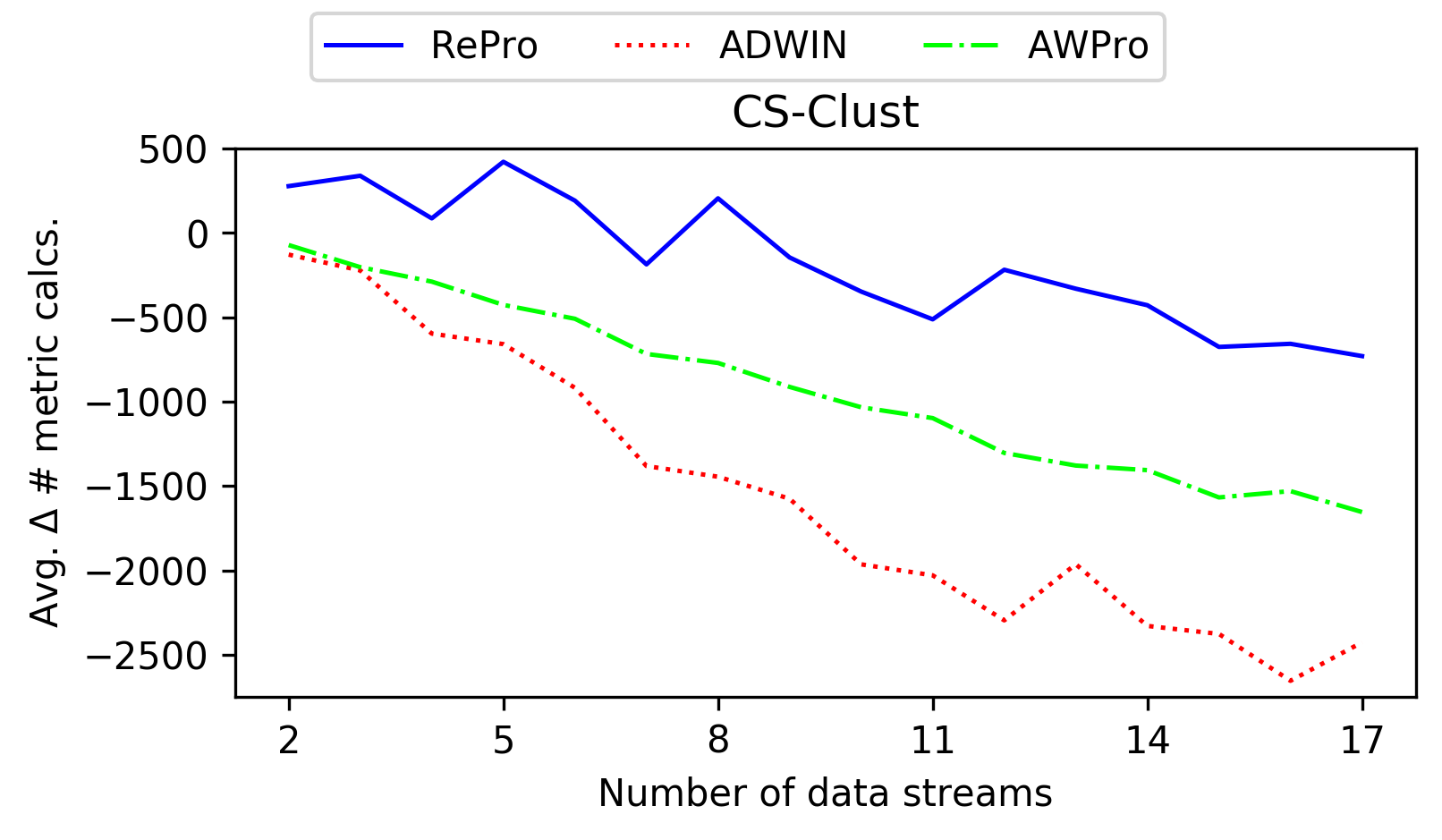}}}
\caption{Change in number of relevancy and diversity metric calculations required to compare and evaluate base models for CS-Clust vs.
    MI-Thresh for increasing numbers of following distance data streams.}
\label{basecomps}
\end{figure}

Although CS-Clust requires additional computation for clustering, the use of conceptual similarity as a measure of
diversity significantly reduces the number of pairwise comparisons between base models. Unlike MI-Thresh,
the diversity metric remains static, independent of concept drifts, and therefore does not need to be recalculated as
the data stream progresses. This is shown in Figure~\ref{basecomps}, which highlights the number of relevancy and
diversity metric calculations
required to compare and evaluate base models in CS-Clust compared to MI-Thresh, as the number of following distance data
streams increases. A significant
reduction in relevancy and diversity metric calculations is observed across all datasets, which can be seen in
Tables~\ref{allSuddenperfs}--\ref{allFollowingperfs}.

\subsection{Local Scaling in STSC}
Parameterless conceptual clustering uses STSC~\cite{Zelnik2005} to create clusters of conceptually similar base models.
Although the number of clusters of base models is determined by STSC, in order to perform SC the affinity matrix used by
STSC undergoes local scaling, as shown in Section~\ref{conceptSimilarity}, using Definition~\ref{def:affinityMatrix}.
This allows better affinities to be obtained
when the density of conceptually similar base models varies~\cite{Zelnik2005}. Throughout this paper a local scaling
parameter of $k=7$ was used, as suggested by Zelnik-Manor and Perona~\cite{Zelnik2005}. Although parameter tuning is not
typically needed for local scaling to perform well~\cite{John2019}, we consider the predictive performance of CS-Clust
and the number of base models used by the OLS meta-learner with local scaling parameters varying between $k=2$ and $k=7$.

\begin{figure*}[tbp]
\centering
\begin{subfigure}[t]{\textwidth}
    \centering
\centerline{\includegraphics[width=0.95\textwidth]{{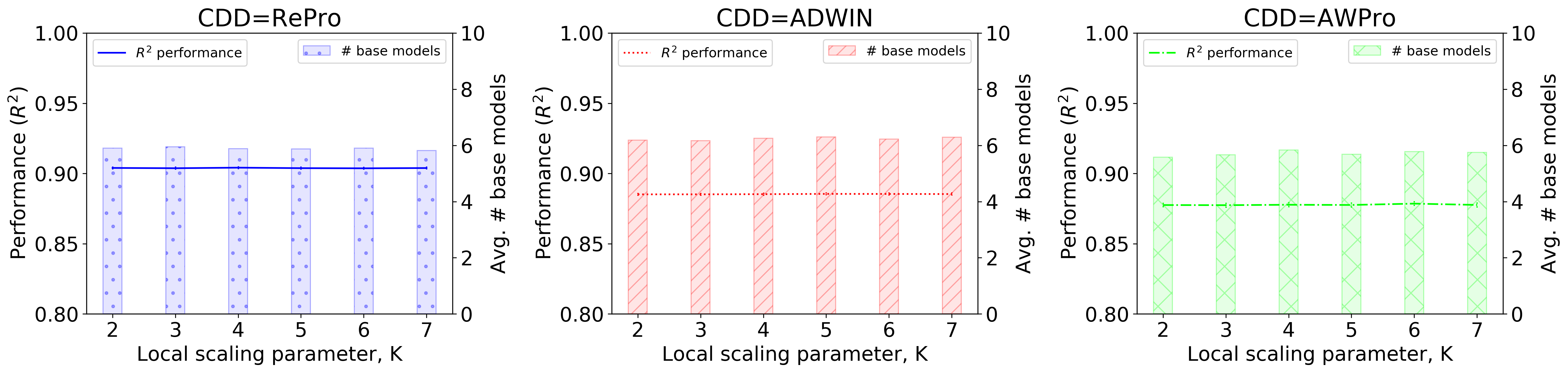}}}
\caption{Sudden drifting hyperplane}
\end{subfigure}

\begin{subfigure}[t]{\textwidth}
    \centering
\centerline{\includegraphics[width=0.95\textwidth]{{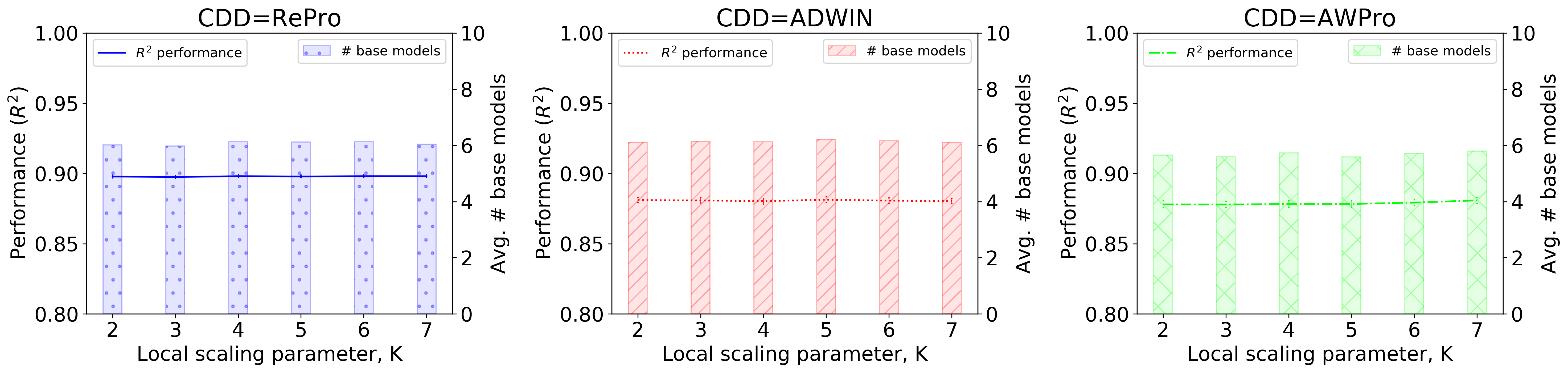}}}
\caption{Gradual drifting hyperplane}
\end{subfigure}
\begin{subfigure}[t]{\textwidth}
    \centering
\centerline{\includegraphics[width=0.95\textwidth]{{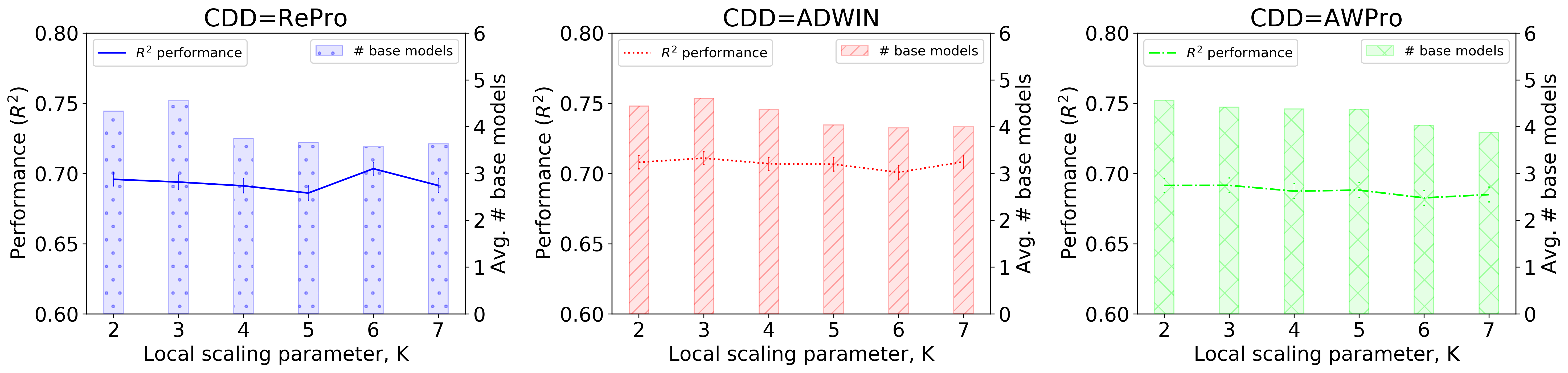}}}
\caption{Heating simulator}
\end{subfigure}
\begin{subfigure}[t]{\textwidth}
    \centering
\centerline{\includegraphics[width=0.95\textwidth]{{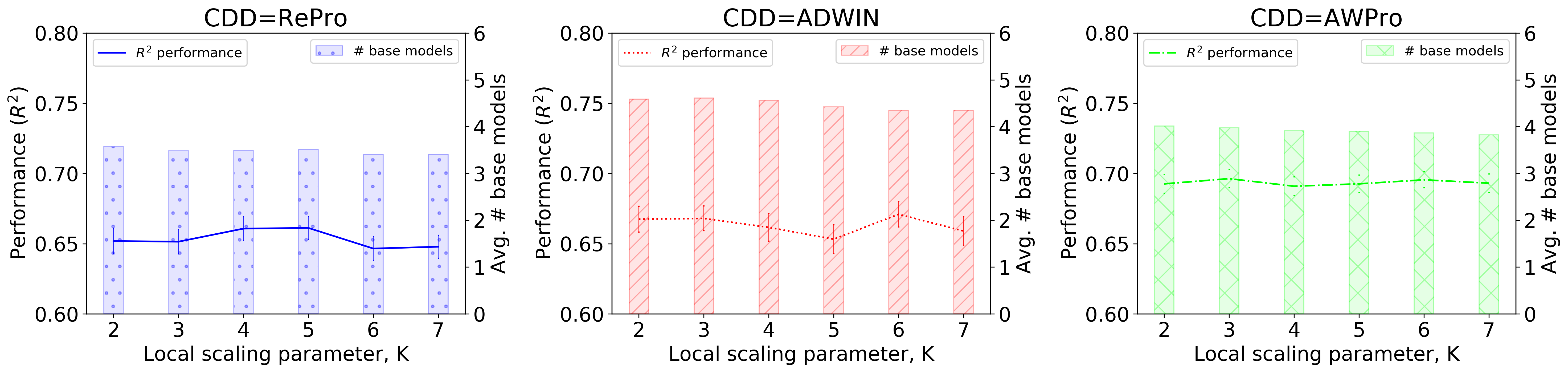}}}
\caption{Following distance}
\end{subfigure}
\caption{$R^{2}$ and number of base models used by CS-Clust meta-learners for varying local scaling parameters in sudden
    drifting hyperplane, gradual drifting hyperplane, heating simulator, and following distance data streams, using
    RePro, ADWIN and AWPro as underlying CDDs.}
\label{localScaling}
\end{figure*}

Figure~\ref{localScaling} presents the predictive performance of CS-Clust and the number of base models selected as
input to the OLS meta-learner using RePro, ADWIN and AWPro as CDDs, for sudden drifting hyperplane, gradual drifting
hyperplane, heating simulator and following distance datasets. Figure~\ref{localScaling} shows that varying the local
scaling parameter, $k$, has minimal effect on the predictive performance of CS-Clust. Additionally, the number of base
models selected as input to the OLS meta-learner does not change significantly. Since CS-Clust selects a single
model from each cluster, the number of base models used by the 
meta-learner is a good indicator of the number of clusters of conceptually similar base models identified.

These results support the suggestions in~\cite{Zelnik2005} and~\cite{John2019} that $k=7$ is a good parameter for local
scaling in STSC. However, accounting for locally dense areas in the affinity matrix can be challenging when clustering base
models learnt from online environments, particularly if CDDs such as ADWIN are used, where base models are not reused in
the presence of recurring concepts. Therefore, future work may consider alternative methods, such as density-aware
kernels~\cite{John2019,Zhang2011} to overcome the challenges of clustering using affinity matrices with locally dense
areas.

\subsection{Reducing Redundant Transfer}
\begin{figure}[tb]
\centering
\centerline{\includegraphics[width=\columnwidth]{{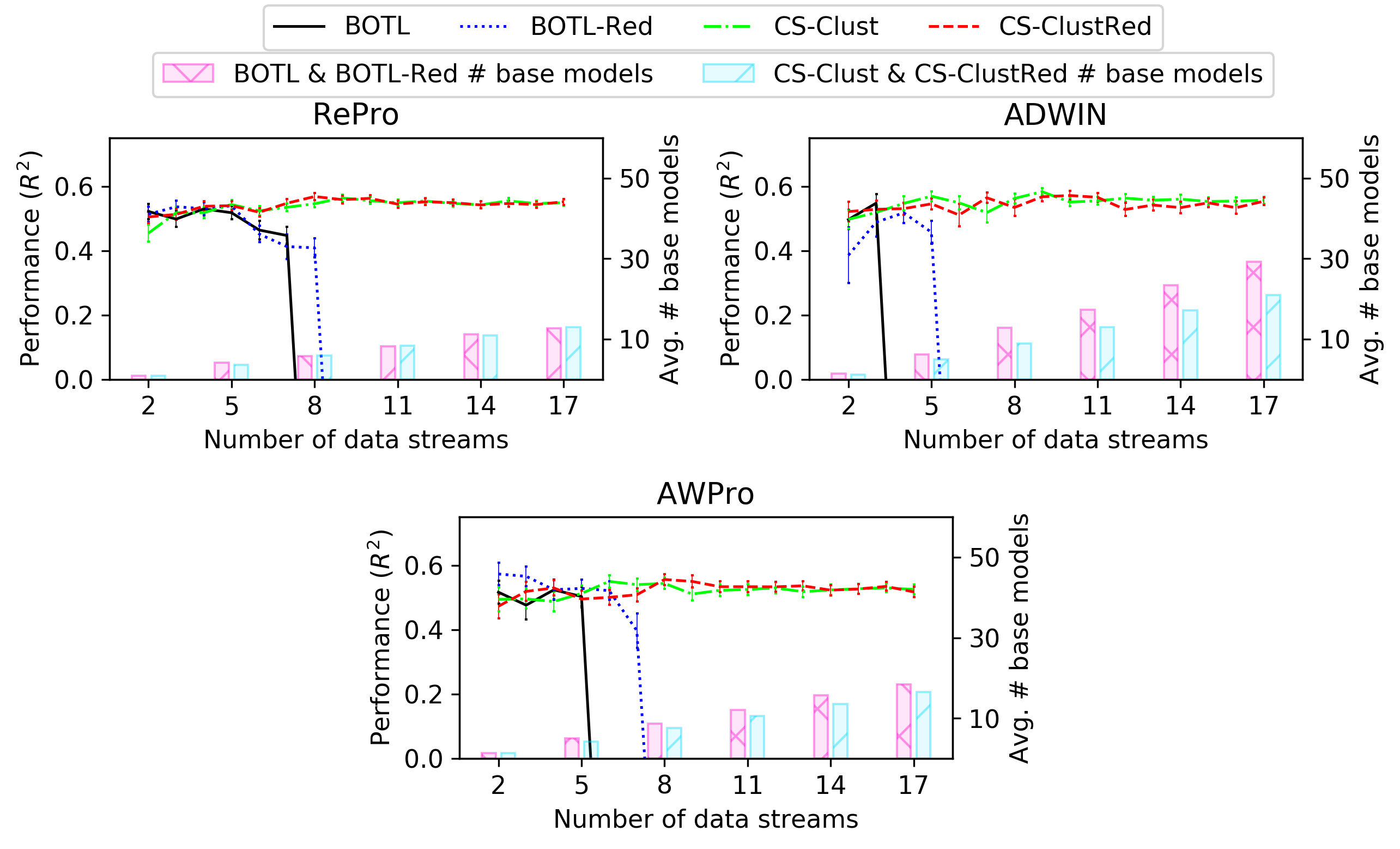}}}
\caption{$R^{2}$ performance, and number of base models used by meta-learners in BOTL, with reduced base model
    transfer (BOTL-Red and CS-ClustRed) compared to transferring every base model (BOTL and CS-Clust) for increasing numbers
of following distance data streams.}
\label{redtransfer}
\end{figure}
To reduce knowledge transfer in BOTL, we used parameterless conceptual clustering on locally learnt base models, prior
to transfer. This allowed
conceptually similar models within a data stream to be identified. In BOTL-Red we apply this to BOTL (with no base
model selection), and in CS-ClustRed to CS-Clust.
Figure~\ref{redtransfer} shows that the number of models transferred can be reduced, without significant change in
predictive performance. ADWIN does not reuse historically learnt models, and therefore a larger reduction in transferred base models
is observed since multiple models learnt from
recurring concepts are no longer transferred. BOTL-Red
does not prevent the meta-learner from overfitting since it only considers the diversity among locally learnt
models, and does not use base model selection techniques to consider the conceptual similarity among models received from
other data streams.

Since CS-ClustRed achieves similar performance to CS-Clust, it may be desirable to use CS-ClustRed when the
communication overhead of knowledge transfer is high. However, there is a trade-off against 
additional computational overhead since conceptual clustering is needed both before knowledge
transfer, to identify conceptually similar locally learnt models, and once base models have been received, to
identify a relevant yet diverse subset of base models to be used by the meta-learner from local and transferred models.

\section{Conclusion}\label{conclusion}
CDDs can be used to make predictions in data streams that are subject to concept
drift, where data availability may be limited.  Meta-learners can improve predictive capabilities
through the use of historically learnt knowledge, or transferred knowledge when online TL is used. However, as the number of
available base models becomes large in comparison to the available data, a relevant yet diverse subset of base models
must be selected to prevent the meta-learner overfitting and to improve generalisation.

We have presented parameterised thresholding and parameterless conceptual clustering methods that estimate conceptual
similarity to
identify diversity among base models, independent of the current distribution in the data stream. Comparable predictive
performance to performance and MI thresholding have been obtained, while requiring fewer base model
comparisons. Although conceptual clustering requires additional computation due to reforming clusters when new
base models are made available, the reduction in base model
comparisons and the avoidance of user defined parameters, makes it more applicable to real-world applications in which the
number of base models and frequency of drifts are unknown and could become large.

Additionally, we have shown that conceptual clustering can be used in online TL to reduce the number of models
transferred between concept drifting data streams. In future work, we will consider cross-domain conceptual similarity
to further reduce the transfer of conceptually similar models. Cross-domain conceptual similarity will also allow locally learnt models to
be replaced by conceptually similar models, learnt from other data streams, that prove to be more effective predictors. As both conceptual similarity and
the BOTL framework are model agnostic, differing machine learning models can be used to create base models within each
data stream, depending on the locally available computational resources. This may be beneficial in online TL frameworks
since a
domain with limited computational capacity can replace locally learnt base models with a conceptually similar base model
that has been learnt in a domain that has greater computational resources. For example, if one domain in a BOTL framework
is only able to utilise simple machine learning algorithms, such as Ridge Regression, to create base models, but other
domains are
capable of using more complex algorithms, such as SVRs, a locally learnt Ridge Regression model
could be replaced by a conceptually similar SVR that has improved predictive capabilities.


\section*{Acknowledgements}
Part of this work was supported by the UK EPSRC and Jaguar Land Rover under the iCASE scheme.

\bibliographystyle{unsrtnat}
\bibliography{bib}  

\begin{thebibliography}{47}
\providecommand{\natexlab}[1]{#1}
\providecommand{\url}[1]{\texttt{#1}}
\expandafter\ifx\csname urlstyle\endcsname\relax
  \providecommand{\doi}[1]{doi: #1}\else
  \providecommand{\doi}{doi: \begingroup \urlstyle{rm}\Url}\fi

\bibitem[Tsymbal(2004)]{Tsymbal2004}
Alexey Tsymbal.
\newblock The problem of concept drift: definitions and related work.
\newblock \emph{Computer Science Department, Trinity College Dublin},
  106\penalty0 (2):\penalty0 58, 2004.

\bibitem[Gama et~al.(2014)Gama, \v{Z}liobait\.{e}, Bifet, Pechenizkiy, and
  Bouchachia]{Gama2014}
Jo{\~a}o Gama, Indr\.{e} \v{Z}liobait\.{e}, Albert Bifet, Mykola Pechenizkiy,
  and Abdelhamid Bouchachia.
\newblock A survey on concept drift adaptation.
\newblock \emph{ACM Comput. Surv.}, 46\penalty0 (4):\penalty0 44:1--44:37,
  2014.
\newblock \doi{10.1145/2523813}.

\bibitem[{\v{Z}}liobait{\.e}(2010)]{Zliobaite2010}
Indr{\.e} {\v{Z}}liobait{\.e}.
\newblock Learning under concept drift: an overview.
\newblock \emph{arXiv preprint arXiv:1010.4784}, 2010.

\bibitem[Bifet and Gavalda(2007)]{Bifet2007}
Albert Bifet and Ricard Gavalda.
\newblock Learning from time-changing data with adaptive windowing.
\newblock In \emph{Proceedings of the 2007 SIAM International Conference on
  Data Mining}, pages 443--448. SIAM, 2007.
\newblock \doi{10.1137/1.9781611972771.42}.

\bibitem[McKay et~al.(2019)McKay, Griffiths, Taylor, Damoulas, and
  Xu]{Mckay2019}
Helen McKay, Nathan Griffiths, Phillip Taylor, Theo Damoulas, and Zhou Xu.
\newblock Online transfer learning for concept drifting data streams.
\newblock In \emph{BigMine@ KDD}, 2019.

\bibitem[McKay et~al.(2020)McKay, Griffiths, Taylor, Damoulas, and
  Xu]{Mckay2020}
Helen McKay, Nathan Griffiths, Phillip Taylor, Theo Damoulas, and Zhou Xu.
\newblock Bi-directional online transfer learning: a framework.
\newblock \emph{Annals of Telecommunications}, 75\penalty0 (9):\penalty0
  523--547, 2020.
\newblock \doi{10.1007/s12243-020-00776-1}.

\bibitem[Brown et~al.(2005)Brown, Wyatt, and Ti{\v{n}}o]{Brown2005}
Gavin Brown, Jeremy~L Wyatt, and Peter Ti{\v{n}}o.
\newblock Managing diversity in regression ensembles.
\newblock \emph{Journal of machine learning research}, 6\penalty0
  (Sep):\penalty0 1621--1650, 2005.

\bibitem[Dutta(2009)]{Dutta2009}
Haimonti Dutta.
\newblock Measuring diversity in regression ensembles.
\newblock In \emph{Proceedings of the 4th Indian International Conference on
  Artificial Intelligence ({IICAI})}, volume~9, pages 2220--2236, 2009.

\bibitem[Brzezinski and Stefanowski(2016)]{Brzezinski2016}
Dariusz Brzezinski and Jerzy Stefanowski.
\newblock Ensemble diversity in evolving data streams.
\newblock In Toon Calders, Michelangelo Ceci, and Donato Malerba, editors,
  \emph{Discovery Science}, pages 229--244, Cham, 2016. Springer International
  Publishing.

\bibitem[Zhuang et~al.(2021)Zhuang, Qi, Duan, Xi, Zhu, Zhu, Xiong, and
  He]{Zhuang2021}
Fuzhen Zhuang, Zhiyuan Qi, Keyu Duan, Dongbo Xi, Yongchun Zhu, Hengshu Zhu, Hui
  Xiong, and Qing He.
\newblock A comprehensive survey on transfer learning.
\newblock \emph{Proceedings of the IEEE}, 109\penalty0 (1):\penalty0 43--76,
  2021.
\newblock \doi{10.1109/JPROC.2020.3004555}.

\bibitem[Zhao et~al.(2014)Zhao, Hoi, Wang, and Li]{Zhao2014}
Peilin Zhao, Steven Hoi, Jialei Wang, and Bin Li.
\newblock Online transfer learning.
\newblock \emph{Artificial Intelligence}, 216:\penalty0 76--102, 2014.

\bibitem[Yang et~al.(2005)Yang, Wu, and Zhu]{Yang2005}
Ying Yang, Xindong Wu, and Xingquan Zhu.
\newblock Combining proactive and reactive predictions for data streams.
\newblock In \emph{Proceedings of the Eleventh ACM SIGKDD International
  Conference on Knowledge Discovery in Data Mining}, KDD '05, pages 710--715.
  ACM, 2005.
\newblock \doi{10.1145/1081870.1081961}.

\bibitem[Koychev(2000)]{Koychev2000}
Ivan Koychev.
\newblock Gradual forgetting for adaptation to concept drift.
\newblock Proceedings of ECAI 2000 Workshop on Current Issues in
  Spatio-Temporal~…, 2000.

\bibitem[Jaber et~al.(2013)Jaber, Cornu{\'e}jols, and Tarroux]{Jaber2013}
Ghazal Jaber, Antoine Cornu{\'e}jols, and Philippe Tarroux.
\newblock Online learning: Searching for the best forgetting strategy under
  concept drift.
\newblock In Minho Lee, Akira Hirose, Zeng-Guang Hou, and Rhee~Man Kil,
  editors, \emph{Neural Information Processing}, pages 400--408, Berlin,
  Heidelberg, 2013. Springer Berlin Heidelberg.

\bibitem[Lu et~al.(2019)Lu, Liu, Dong, Gu, Gama, and Zhang]{Lu2019}
Jie Lu, Anjin Liu, Fan Dong, Feng Gu, João Gama, and Guangquan Zhang.
\newblock Learning under concept drift: A review.
\newblock \emph{IEEE Transactions on Knowledge and Data Engineering},
  31\penalty0 (12):\penalty0 2346--2363, 2019.
\newblock \doi{10.1109/TKDE.2018.2876857}.

\bibitem[Yang et~al.(2006)Yang, Wu, and Zhu]{Yang2006}
Ying Yang, Xindong Wu, and Xingquan Zhu.
\newblock Mining in anticipation for concept change: Proactive-reactive
  prediction in data streams.
\newblock \emph{Data Mining and Knowledge Discovery}, 13\penalty0 (3):\penalty0
  261--289, 2006.
\newblock \doi{10.1145/1081870.1081961}.

\bibitem[{Ramamurthy} and {Bhatnagar}(2007)]{Ramamurthy2007}
S.~{Ramamurthy} and R.~{Bhatnagar}.
\newblock Tracking recurrent concept drift in streaming data using ensemble
  classifiers.
\newblock In \emph{Sixth International Conference on Machine Learning and
  Applications (ICMLA 2007)}, pages 404--409, Dec 2007.
\newblock \doi{10.1109/ICMLA.2007.80}.

\bibitem[Bifet(2009)]{BifetThesis}
Albert Bifet.
\newblock Adaptive learning and mining for data streams and frequent patterns.
\newblock \emph{SIGKDD Explor. Newsl.}, 11\penalty0 (1):\penalty0 55--56,
  November 2009.
\newblock ISSN 1931-0145.
\newblock \doi{10.1145/1656274.1656287}.

\bibitem[Bifet et~al.(2009)Bifet, Holmes, Pfahringer, Kirkby, and
  Gavald\`{a}]{Bifet2009}
Albert Bifet, Geoff Holmes, Bernhard Pfahringer, Richard Kirkby, and Ricard
  Gavald\`{a}.
\newblock New ensemble methods for evolving data streams.
\newblock In \emph{Proceedings of the 15th ACM SIGKDD International Conference
  on Knowledge Discovery and Data Mining}, KDD '09, pages 139--148, New York,
  NY, USA, 2009. Association for Computing Machinery.
\newblock ISBN 9781605584959.
\newblock \doi{10.1145/1557019.1557041}.

\bibitem[Du and Swamy(2019)]{Du2019}
Ke-Lin Du and M.~N.~S. Swamy.
\newblock \emph{Combining Multiple Learners: Data Fusion and Ensemble
  Learning}, pages 737--767.
\newblock Springer London, London, 2019.
\newblock ISBN 978-1-4471-7452-3.
\newblock \doi{10.1007/978-1-4471-7452-3_25}.

\bibitem[Gomes et~al.(2017)Gomes, Barddal, Enembreck, and Bifet]{Gomes2017}
Heitor~Murilo Gomes, Jean~Paul Barddal, Fabr\'{\i}cio Enembreck, and Albert
  Bifet.
\newblock A survey on ensemble learning for data stream classification.
\newblock \emph{ACM Comput. Surv.}, 50\penalty0 (2), March 2017.
\newblock ISSN 0360-0300.
\newblock \doi{10.1145/3054925}.

\bibitem[Kolter and Maloof(2007)]{Kolter2007}
J~Zico Kolter and Marcus~A Maloof.
\newblock Dynamic weighted majority: An ensemble method for drifting concepts.
\newblock \emph{Journal of Machine Learning Research}, 8\penalty0
  (Dec):\penalty0 2755--2790, 2007.

\bibitem[{Gomes Soares} and Araújo(2015)]{GomesSoares2015}
Symone {Gomes Soares} and Rui Araújo.
\newblock An online weighted ensemble of regressor models to handle concept
  drifts.
\newblock \emph{Engineering Applications of Artificial Intelligence},
  37:\penalty0 392--406, 2015.
\newblock ISSN 0952-1976.
\newblock \doi{10.1016/j.engappai.2014.10.003}.

\bibitem[Kolter and Maloof(2005)]{Kolter2005}
Jeremy~Z. Kolter and Marcus~A. Maloof.
\newblock Using additive expert ensembles to cope with concept drift.
\newblock In \emph{Proceedings of the 22nd International Conference on Machine
  Learning}, ICML '05, pages 449--456. ACM, 2005.
\newblock \doi{10.1145/1102351.1102408}.

\bibitem[Ge et~al.(2013)Ge, Gao, and Zhang]{Ge2013}
Liang Ge, Jing Gao, and Aidong Zhang.
\newblock Oms-tl: A framework of online multiple source transfer learning.
\newblock In \emph{Proceedings of the 22nd ACM International Conference on
  Information \& Knowledge Management}, CIKM '13, pages 2423--2428, New York,
  NY, USA, 2013. Association for Computing Machinery.
\newblock ISBN 9781450322638.
\newblock \doi{10.1145/2505515.2505603}.
\newblock URL \url{https://doi.org/10.1145/2505515.2505603}.

\bibitem[Grubinger et~al.(2016)Grubinger, Chasparis, and
  Natschl{\"a}ger]{Grubinger2016GOTL}
Thomas Grubinger, Georgios Chasparis, and Thomas Natschl{\"a}ger.
\newblock Online transfer learning for climate control in residential
  buildings.
\newblock In \emph{Proceedings of the 5th Annual European Control Conference
  (ECC 2016)}, pages 1183--1188, 2016.
\newblock \doi{10.1109/ECC.2016.7810450}.

\bibitem[Wu et~al.(2017)Wu, Zhou, Yan, Wu, and Min]{Wu2017}
Qingyao Wu, Xiaoming Zhou, Yuguang Yan, Hanrui Wu, and Huaqing Min.
\newblock Online transfer learning by leveraging multiple source domains.
\newblock \emph{Knowledge and Information Systems}, 52\penalty0 (3):\penalty0
  687--707, 2017.

\bibitem[Wang et~al.(2003)Wang, Fan, Yu, and Han]{Wang2003}
Haixun Wang, Wei Fan, Philip~S. Yu, and Jiawei Han.
\newblock Mining concept-drifting data streams using ensemble classifiers.
\newblock In \emph{Proceedings of the Ninth ACM SIGKDD International Conference
  on Knowledge Discovery and Data Mining}, KDD '03, pages 226--235, New York,
  NY, USA, 2003. Association for Computing Machinery.
\newblock ISBN 1581137370.
\newblock \doi{10.1145/956750.956778}.

\bibitem[Rokach(2010)]{Rokach2010}
Lior Rokach.
\newblock Ensemble-based classifiers.
\newblock \emph{Artificial intelligence review}, 33\penalty0 (1):\penalty0
  1--39, 2010.
\newblock \doi{10.1007/s10462-009-9124-7}.

\bibitem[Vapnik(2013)]{Vapnik1995}
Vladimir Vapnik.
\newblock \emph{The nature of statistical learning theory}.
\newblock Springer science \& business media, 2013.

\bibitem[Friedman(1997)]{Friedman1997}
Jerome~H. Friedman.
\newblock On bias, variance, 0/1---loss, and the curse-of-dimensionality.
\newblock \emph{Data Mining and Knowledge Discovery}, 1\penalty0 (1):\penalty0
  55--77, Mar 1997.
\newblock ISSN 1573-756X.
\newblock \doi{10.1023/A:1009778005914}.

\bibitem[Zhou(2012)]{Zhou2012}
Zhi-Hua Zhou.
\newblock \emph{Ensemble methods: foundations and algorithms}.
\newblock Chapman and Hall/CRC, 2012.

\bibitem[Cherkassky and Mulier(2007)]{Cherkassky2007}
Vladimir Cherkassky and Filip~M Mulier.
\newblock \emph{Learning from data: concepts, theory, and methods}.
\newblock John Wiley \& Sons, 2007.

\bibitem[{Niyogi} and {Girosi}(1996)]{Niyogi1994}
P.~{Niyogi} and F.~{Girosi}.
\newblock On the relationship between generalization error, hypothesis
  complexity, and sample complexity for radial basis functions.
\newblock \emph{Neural Computation}, 8\penalty0 (4):\penalty0 819--842, 1996.
\newblock \doi{10.1162/neco.1996.8.4.819}.

\bibitem[Bj{\"o}rck and Golub(1973)]{Bjorck1973}
{\.{A}}ke Bj{\"o}rck and Gene~H Golub.
\newblock Numerical methods for computing angles between linear subspaces.
\newblock \emph{Mathematics of computation}, 27\penalty0 (123):\penalty0
  579--594, 1973.
\newblock \doi{10.2307/2005662}.

\bibitem[Golub and Zha(1995)]{Golub1995}
Gene~H. Golub and Hongyuan Zha.
\newblock The canonical correlations of matrix pairs and their numerical
  computation.
\newblock In Adam Bojanczyk and George Cybenko, editors, \emph{Linear Algebra
  for Signal Processing}, pages 27--49, New York, NY, 1995. Springer New York.
\newblock ISBN 978-1-4612-4228-4.

\bibitem[Knyazev and Zhu(2012)]{Knyazev2012}
Andrew~V Knyazev and Peizhen Zhu.
\newblock Principal angles between subspaces and their tangents.
\newblock \emph{arXiv preprint arXiv:1209.0523}, 2012.

\bibitem[Zhao et~al.(2007)Zhao, Lin, and Tang]{Zhao2007}
Deli Zhao, Zhouchen Lin, and Xiaoou Tang.
\newblock Laplacian pca and its applications.
\newblock In \emph{2007 IEEE 11th International Conference on Computer Vision},
  pages 1--8, 2007.
\newblock \doi{10.1109/ICCV.2007.4409096}.

\bibitem[Xu et~al.(2010)Xu, Caramanis, and Sanghavi]{Xu2010}
Huan Xu, Constantine Caramanis, and Sujay Sanghavi.
\newblock Robust pca via outlier pursuit.
\newblock \emph{arXiv preprint arXiv:1010.4237}, 2010.

\bibitem[John et~al.(2019)John, Watson, Barnes, Pitzalis, and Lewis]{John2019}
Christopher~R John, David Watson, Michael~R Barnes, Costantino Pitzalis, and
  Myles~J Lewis.
\newblock {Spectrum: fast density-aware spectral clustering for single and
  multi-omic data}.
\newblock \emph{Bioinformatics}, 36\penalty0 (4):\penalty0 1159--1166, 09 2019.
\newblock ISSN 1367-4803.
\newblock \doi{10.1093/bioinformatics/btz704}.
\newblock URL \url{https://doi.org/10.1093/bioinformatics/btz704}.

\bibitem[Zelnik-Manor and Perona(2004)]{Zelnik2005}
Lihi Zelnik-Manor and Pietro Perona.
\newblock Self-tuning spectral clustering.
\newblock In \emph{Proceedings of the 17th International Conference on Neural
  Information Processing Systems}, NIPS'04, pages 1601--1608, Cambridge, MA,
  USA, 2004. MIT Press.
\newblock \doi{10.5555/2976040.2976241}.

\bibitem[Zhang et~al.(2011)Zhang, Li, and Yu]{Zhang2011}
Xianchao Zhang, Jingwei Li, and Hong Yu.
\newblock Local density adaptive similarity measurement for spectral
  clustering.
\newblock \emph{Pattern Recognition Letters}, 32\penalty0 (2):\penalty0
  352--358, 2011.
\newblock ISSN 0167-8655.
\newblock \doi{https://doi.org/10.1016/j.patrec.2010.09.014}.
\newblock URL
  \url{https://www.sciencedirect.com/science/article/pii/S0167865510003181}.

\bibitem[Hammerla and Pl\"{o}tz(2015)]{Hammerla2015}
Nils~Y. Hammerla and Thomas Pl\"{o}tz.
\newblock Let’s (not) stick together: Pairwise similarity biases
  cross-validation in activity recognition.
\newblock In \emph{Proceedings of the 2015 ACM International Joint Conference
  on Pervasive and Ubiquitous Computing}, UbiComp ’15, pages 1041--1051.
  Association for Computing Machinery, 2015.
\newblock ISBN 9781450335744.
\newblock \doi{10.1145/2750858.2807551}.

\bibitem[Yang et~al.(2019)Yang, Liu, Nie, and Liu]{Yang2019}
Libo Yang, Xuemei Liu, Feiping Nie, and Mingtang Liu.
\newblock Large-scale spectral clustering based on representative points.
\newblock \emph{Mathematical Problems in Engineering}, 2019, 2019.

\bibitem[Kira and Rendell(1992)]{Kira1992}
Kenji Kira and Larry~A. Rendell.
\newblock A practical approach to feature selection.
\newblock In Derek Sleeman and Peter Edwards, editors, \emph{Machine Learning
  Proceedings 1992}, pages 249--256. Morgan Kaufmann, San Francisco (CA), 1992.
\newblock ISBN 978-1-55860-247-2.
\newblock \doi{10.1016/B978-1-55860-247-2.50037-1}.

\bibitem[{Ueda} and {Nakano}(1996)]{Ueda1996}
N.~{Ueda} and R.~{Nakano}.
\newblock Generalization error of ensemble estimators.
\newblock In \emph{Proceedings of International Conference on Neural Networks
  (ICNN'96)}, volume~1, pages 90--95 vol.1, June 1996.
\newblock \doi{10.1109/ICNN.1996.548872}.

\bibitem[Ren et~al.(2016)Ren, Zhang, and Suganthan]{Ren2016}
Ye~Ren, Le~Zhang, and P.N. Suganthan.
\newblock Ensemble classification and regression-recent developments,
  applications and future directions [review article].
\newblock \emph{IEEE Computational Intelligence Magazine}, 11\penalty0
  (1):\penalty0 41--53, 2016.
\newblock \doi{10.1109/MCI.2015.2471235}.

\end{thebibliography}

\end{document}